\documentclass{article}
\usepackage{arxiv}
\usepackage[utf8]{inputenc} 
\usepackage[T1]{fontenc}    
\usepackage{hyperref}       
\usepackage{url}            
\usepackage{booktabs}       
\usepackage{amsfonts}       
\usepackage{amsmath}
\usepackage{nicefrac}       
\usepackage{microtype}      
\usepackage{cleveref}       
\newcommand{\headeright}{}  
\usepackage{graphicx}
\usepackage{natbib}
\usepackage{doi}
\def\onedot{.\spacefactor3000 }
\def\etal{\emph{et al}\onedot}
\usepackage{pifont}
\newcommand{\cmark}{\textcolor{green!60!black}{\ding{51}}} 
\newcommand{\xmark}{\textcolor{red!70!black}{\ding{55}}}   
\usepackage[table]{xcolor}
\newcommand{\mysection}[1]
{\vspace{2pt}\noindent\textbf{#1}}

\usepackage{tikz}
\usepackage{array}
\newcolumntype{C}[1]{>{\centering\arraybackslash}m{#1}}
\usepackage[ruled, vlined,noend, linesnumbered, commentsnumbered]{algorithm2e}
\RequirePackage[font=footnotesize,skip=3pt,subrefformat=parens]{subcaption}

\definecolor{commentsColor}{RGB}{219, 48, 122}

\SetCommentSty{mycommfont}
\let\oldnl\nl
\newcommand{\nonl}{\renewcommand{\nl}{\let\nl\oldnl}}
\definecolor{commentcolor}{RGB}{140,140,140}
\newcommand{\PyComment}[1]{\fontfamily{pcr}\selectfont\textcolor{commentcolor}{\# #1}}
\newcommand{\PyCode}[1]{\fontfamily{pcr}\selectfont #1}

\title{CAPTAIN: Semantic Feature Injection for Memorization Mitigation in Text-to-Image Diffusion Models}

\date{}

\author{\href{https://zhangt-tech.github.io/}{\hspace{1mm}Tong Zhang}, \href{http://carloshinojosa.me/}{\hspace{1mm} Carlos Hinojosa}, \href{https://www.bernardghanem.com/}{\hspace{1mm}Bernard Ghanem} \\
\texttt{\{tong.zhang.1; carlos.hinojosa; bernard.ghanem\}@kaust.edu.sa}
\\
	King Abdullah University of Science and Technology
}

\renewcommand{\headeright}{}
\renewcommand{\undertitle}{}
\renewcommand{\shorttitle}{}

\begin{document}
\maketitle


\begin{abstract}
Diffusion models can unintentionally reproduce training examples, raising privacy and copyright concerns as these systems are increasingly deployed at scale. Existing inference-time mitigation methods typically manipulate classifier-free guidance (CFG) or perturb prompt embeddings; however, they often struggle to reduce memorization without compromising alignment with the conditioning prompt. We introduce CAPTAIN, a training-free framework that mitigates memorization by directly modifying latent features during denoising. CAPTAIN first applies frequency-based noise initialization to reduce the tendency to replicate memorized patterns early in the denoising process. It then identifies the optimal denoising timesteps for feature injection and localizes memorized regions. Finally, CAPTAIN injects semantically aligned features from non-memorized reference images into localized latent regions, suppressing memorization while preserving prompt fidelity and visual quality. Our experiments show that CAPTAIN achieves substantial reductions in memorization compared to CFG-based baselines while maintaining strong alignment with the intended prompt.
\end{abstract}    
\section{Introduction}
\label{sec:introduction}

Diffusion models have made significant progress in text-to-image synthesis, generating high-quality, diverse, and stylistically rich images. However, recent studies have revealed a concerning trend: some of these ``novel'' creations are, in fact, near-exact reproductions of images from their training
datasets, a behavior known as \textit{memorization}~\cite{somepalli2023diffusion, carlini2023extracting, somepalli2023understanding, gu2025on}. This unintended memorization raises serious concerns for both model owners and end users, particularly when training data contain sensitive or copyrighted material, and as large-scale diffusion systems are increasingly deployed in public and commercial settings. A recent example is the UK High Court case involving Getty Images and Stability AI, in which generated samples from Stable Diffusion were found to reproduce Getty's watermark, highlighting the real-world risks of memorization and copyright infringement~\cite{guardian2025getty}. Understanding and mitigating memorization has therefore become crucial for ensuring the safe and responsible use of generative models.

\begin{figure}[t]
    \centering
    \includegraphics[width=0.9\columnwidth]{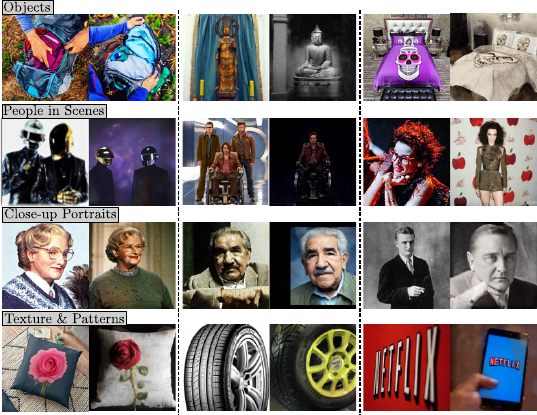}
    \caption{CAPTAIN achieves strong semantic alignment and effective memorization mitigation across diverse categories, including Objects, People in Scenes, Close-up Portraits (the most challenging category), and Textures \& Patterns, where structural repetition is more likely. In each pair, the left image shows the memorized training example, while the right image shows the corresponding mitigated result generated by our approach.}
    \label{fig:teaser}
\end{figure}

Existing mitigation strategies can be broadly categorized into training-time and inference-time approaches. Training-time methods typically require access to the original dataset and large computational resources, limiting their practicality~\cite{wen2024detecting}. In contrast, inference-time methods aim to detect or suppress memorization without modifying model parameters~\cite{chen2024towards, jain2025classifier, hintersdorf2024finding, chen2025exploring, ren2024unveiling}, offering an efficient alternative.

Overall, most inference-time strategies steer the diffusion process away from memorized outputs by manipulating the strength of classifier-free guidance (CFG), prompt embeddings, or cross-attention behavior. For instance,~\cite{wen2024detecting} detect memorization by monitoring the magnitude of the conditional noise prediction and mitigate it by adjusting prompt embeddings.~\cite{jain2025classifier} propose applying opposite guidance during early denoising steps to counteract memorization induced by CFG. Similarly, \cite{chen2025enhancing} mitigate memorization by re-anchoring prompts within CFG through prompt perturbation and semantic prompt search for non-memorized variants. On the other hand, \cite{chen2025exploring} identify a novel ``bright ending'' (BE) anomaly in text-to-image diffusion models, where memorized image patches exhibit significantly greater attention to the final text token during the last inference step than non-memorized patches. This distinct cross-attention pattern highlights regions where the generated image replicates training data, enabling efficient localization of memorized regions. However, these methods often struggle to reduce memorization without compromising alignment with the conditioning prompt.

In this paper, we introduce CAPTAIN (see ~\cref{fig:teaser}), a training-free framework that mitigates memorization at inference time. Instead of manipulating CFG or perturbing prompt embeddings to control generation, we guide the denoising trajectory away from memorized outputs by directly modifying latent features. The contributions are as follows:
\begin{itemize}
    \item[(i)] \textbf{Frequency decomposed initialization.} Our initialization strategy blends the low-frequency components of a given reference image with the high-frequency components of random noise. This constrains the initial latent and encourages the early denoising trajectory to steer away from reproducing memorized patterns.

    \item[(ii)] \textbf{Timestep and spatial memorization localization.} Our method operates directly in the latent space and introduces a timestep window localization strategy that determines when feature injection is most effective during denoising and identifies the spatial regions that contain memorized content. This enables targeted suppression of memorized content while preserving the semantic structure of the generated image.

    \item[(iii)] \textbf{Semantic feature injection for memorization mitigation.} We propose a localized feature injection mechanism that replaces memorized latent regions with semantically aligned features extracted from non-memorized reference images, steering generation away from memorized outputs while maintaining prompt fidelity and visual quality.
\end{itemize}

\section{Related Works}
\label{sec:related_works}

\mysection{Memorization in Diffusion Models.} Diffusion models can unintentionally reproduce images that closely resemble samples seen during training, a behavior referred to as \textit{memorization}. This phenomenon raises privacy and copyright concerns, as it can lead to the leakage of sensitive or proprietary data. Recent studies have analyzed this issue~\cite{somepalli2023diffusion, carlini2023extracting}, showing that large pretrained text-to-image models such as Stable Diffusion can directly replicate training images under specific prompts. Although newer versions (e.g., SDv2.1) apply dataset de-duplication, subsequent research demonstrates that memorization persists and cannot be fully explained by data duplication alone ~\cite{somepalli2023understanding, gu2025on}. Several contributing factors have been identified, including model capacity, dataset scale and diversity, and prompt conditioning mechanisms. Together, these findings suggest that memorization arises not only from data redundancy but also from how diffusion models encode and retrieve semantic and structural priors during denoising, motivating the development of inference-time mitigation strategies beyond dataset curation or retraining.

\mysection{Detection and Mitigation Strategies.} Early efforts to address memorization in diffusion models operated either during training or at inference. Training-time methods typically require access to the original dataset and large computational resources, limiting their practicality. As a result, recent studies have focused on inference-time detection and mitigation~\cite{chen2024towards, jain2025classifier, hintersdorf2024finding, chen2025exploring, ren2024unveiling}. ~\cite{wen2024detecting} detect memorization by monitoring the magnitude of conditional noise prediction and mitigate it by adjusting prompt embeddings. ~\cite{chen2025exploring} further refine this process through the Bright Ending (BE) attention mechanism, which identifies memorized regions where the last cross-attention layer assigns abnormally high attention to the final prompt token. ~\cite{somepalli2023understanding} alleviate memorization by perturbing input prompts via token addition or replacement, while 
~\cite{ren2024unveiling} link memorization to highly concentrated cross-attention patterns, where specific tokens act as ``triggers''.~\cite{jain2025classifier} propose applying opposite guidance during early denoising steps to counteract memorization induced by classifier-free guidance (CFG). Similarly, ~\cite{hanadjusting} adjust the initial noise samples to reduce memorization. More recently, ~\cite{chen2025enhancing} introduced PRSS, which mitigates memorization by re-anchoring prompts within CFG through prompt perturbation and semantic prompt search for non-memorized variants. Overall, most inference-time strategies steer the diffusion process away from memorized outputs by manipulating CFG strength, prompt embeddings, or cross-attention behavior. In contrast, our proposed method, CAPTAIN, mitigates memorization by directly operating in the latent feature space, combining frequency-aware initialization and semantic feature injection to steer denoising away from memorized content without modifying CFG or prompt embeddings.

\mysection{Feature Injection in Diffusion Models} Feature-space interventions in diffusion models have been mainly explored for image editing and controllability rather than memorization mitigation. Methods such as Plug-and-Play Diffusion~\cite{tumanyan2023plug} and Prompt-to-Prompt~\cite{hertz2023prompttoprompt} manipulate cross-attention features to control spatial layouts or preserve structure during prompt editing, while Null-Text Inversion~\cite{mokady2023null} and SDEdit~\cite{meng2022sdedit} guide generation by modifying latent representations. As shown in the video domain, FreeInit~\cite{wu2024freeinit} performs frequency-domain filtering to inject low-frequency structural information from an anchor frame, thereby enabling long-term temporal stability in videos. More recently, MoCA-Video~\cite{zhang2025motion} extends concept-level alignment to the video domain by leveraging motion-aware semantic feature fusion, ensuring temporal consistency across edited sequences. In contrast to these approaches, which inject or modify features to improve editability and temporal coherence, our method initializes with non-memorized content and  explicitly injects localized features to further suppress memorized content, integrating semantic cues to preserve concept fidelity while preventing overfitting to the training data.



\begin{figure*}[t]
    \centering
    \includegraphics[width=\linewidth]{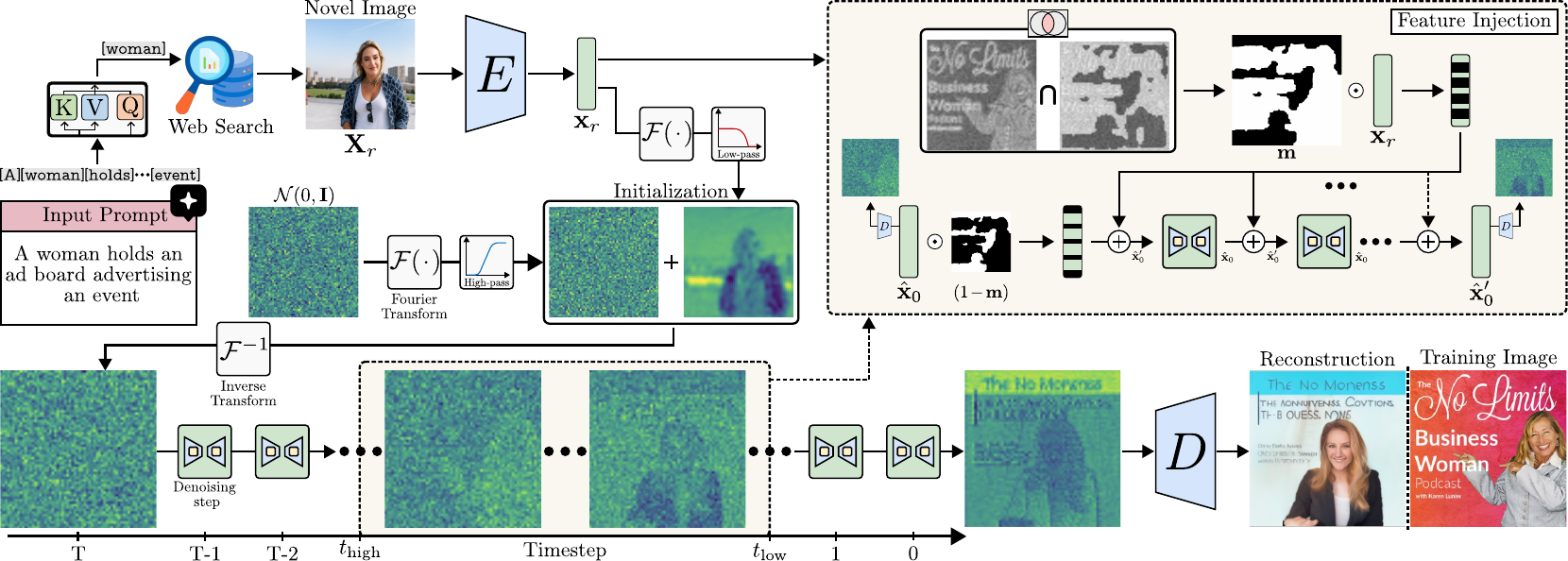}
    \caption{Given an input prompt, CAPTAIN retrieves a semantically related but unseen reference image from the web and encodes it into latent features $\mathbf{x}_r$. (Left) We initialize the diffusion process using a frequency-based strategy: the low-frequency components of Gaussian noise are combined with the high-frequency components of the reference image to discourage early-stage memorization. (Bottom) During denoising, we monitor image–text alignment to identify the optimal injection window $t \in [t_{\mathrm{high}}, t_{\mathrm{low}}]$. (Right) By intersecting the BE mask with concept-specific attention maps, we produce a binary mask $\mathbf{m}$ that highlights the target memorized regions. At every step $t \in [t_{\mathrm{high}}, t_{\mathrm{low}}]$, CAPTAIN injects semantically aligned features from $\mathbf{x}_r$ into the masked regions of $\hat{\mathbf{x}}_0$, yielding the updated latent $\hat{\mathbf{x}}'_0$.}
    \label{fig:pipeline}
\end{figure*}
\section{Method}
\label{sec:method}

To address memorization in text-to-image diffusion models, we propose CAPTAIN, a training-free framework that steers the denoising trajectory toward non-memorized outputs by injecting features directly into the latent denoising process with frequency decomposed  noise initialization. The framework integrates detection and mitigation within a unified pipeline that adaptively determines the semantic content, spatial regions, and intermediate timesteps for feature injection, effectively mitigating memorization while maintaining prompt consistency.

\mysection{Framework Overview.} Our framework is illustrated in \cref{fig:pipeline}. Given a text prompt $\mathbf{p}$ and a reference non-memorized image $\mathbf{X}_{r}$ (see \cref{sec:init}), the goal is to generate an image that remains semantically aligned with the prompt while avoiding the reproduction of memorized content from the training data. CAPTAIN operates through four stages. First, we initialize the latent variable by combining the low-frequency components of $\mathbf{X}_{r}$ with the high-frequency components of Gaussian noise $\epsilon_T \sim \mathcal{N}(0, \mathbf{I})$ (\cref{sec:init}), discouraging structural copying from training samples. During denoising, we hypothesize that memorization tends to appear after coarse structure formation but before fine-detail refinement; therefore, we identify a timestep window where feature injection can effectively steer the generation toward a non-memorized output (\cref{sec:TML}). Once this window is determined, we localize potential memorized regions by combining Bright Ending (BE) attention with concept-specific attention maps, isolating areas where memorization overlaps with the target semantic concept (\cref{sec:SML}). Finally, within those localized regions, we inject semantically consistent but visually distinct features from conditioned image, guiding the denoising trajectory toward outputs that maintain prompt fidelity while reducing the influence of memorized content (\cref{sec:mitigation}).

\mysection{Preliminaries on Diffusion Models.} Diffusion probabilistic models~\cite{ho2020denoising} define a forward–reverse process that transforms clean data samples $\mathbf{x}_0$ into Gaussian noise through a Markov chain and learns to invert this noising process. In latent diffusion models~\cite{rombach2022high}, this process operates in a compressed latent space rather than in pixel space; an encoder $E$ first maps images to the latent domain where diffusion occurs, and a decoder $D$ later reconstructs them. In this work, $\mathbf{x}_t$ denotes a latent variable that evolves entirely in this latent space, with $\mathbf{x}_0$ representing the final latent after denoising. The forward diffusion gradually perturbs $\mathbf{x}_0$ according to a variance schedule $\{\beta_t\}_{t=1}^T$:
\begin{equation}
q(\mathbf{x}_t \mid \mathbf{x}_{t-1}) = \mathcal{N}\!\left(\sqrt{1-\beta_t}\,\mathbf{x}_{t-1},\, \beta_t \mathbf{I}\right),
\label{eq:forward}
\end{equation}
so that after $T$ steps, $\mathbf{x}_T \sim \mathcal{N}(0, \mathbf{I})$. The reverse process learns to denoise $\mathbf{x}_t$ step by step using a neural network parameterized by $\theta$:
\begin{equation}
p_\theta(\mathbf{x}_{t-1}\mid \mathbf{x}_t) = 
\mathcal{N}\!\left(\mu_\theta(\mathbf{x}_t,t,c),\sigma_t^2\mathbf{I}\right),
\label{eq:reverse}
\end{equation}
where $c$ denotes a conditioning input. Following \cite{ho2020denoising, song2021denoising, rombach2022high}, the mean term can be expressed in closed form through the predicted noise $\epsilon_\theta(\mathbf{x}_t,t,c)$:
\begin{equation}
    \mu_\theta(\mathbf{x}_t, t, c)
=\frac{1}{\sqrt{\alpha_t}}\!\left(\mathbf{x}_t
-\frac{1 - \alpha_t}{\sqrt{1 - \bar{\alpha}_t}}\,\epsilon_\theta(\mathbf{x}_t, t, c)\right),
\end{equation}
where $\alpha_t = 1 - \beta_t$ and $\bar{\alpha}_t = \prod_{s=1}^{t}\alpha_s$. The denoising network $\epsilon_\theta$, typically a U-Net~\cite{rombach2022high}, is trained by minimizing the simplified objective
\begin{equation}
\mathcal{L}_{\text{simple}} = 
\mathbb{E}_{t,\mathbf{x}_0,\epsilon}\!\left[\|\epsilon - \epsilon_\theta(\mathbf{x}_t, t, c)\|_2^2\right],
\label{eq:loss}
\end{equation}
where $\epsilon \sim \mathcal{N}(0, \mathbf{I})$ is standard Gaussian noise. At inference, samples are generated by iteratively predicting the clean latent
\begin{equation}
    \hat{\mathbf{x}}_0 = \frac{\mathbf{x}_t - \sqrt{1-\bar{\alpha}_t} \epsilon_\theta (\mathbf{x}_t, t, c)}{\sqrt{\bar{\alpha}_t}},
    \label{eq:predicted_x0}
\end{equation}
and reconstructing the previous step:
\begin{equation}
 \mathbf{x}_{t-1} = \sqrt{\bar{\alpha}_{t-1}} \hat{\mathbf{x}}_0 + \sqrt{1-\bar{\alpha}_{t-1}} \epsilon_\theta (\mathbf{x}_t, t, c) + \sigma_t \epsilon_t,
 \label{eq:next_step_latent}
\end{equation}
where $\epsilon_t \sim \mathcal{N}(0, \mathbf{I})$ and $\sigma_t$ follows the chosen noise schedule. Finally, a decoder $D$ reconstructs the generated image $\mathbf{X} = D(\mathbf{x}_0)$ from the latent space, enabling efficient high-resolution synthesis while preserving semantic structure.

\subsection{Initialization and Reference Image Selection}
\label{sec:init}

\mysection{Initialization via Frequency Decomposition.} Drawing inspiration from recent works demonstrating that initialization influences memorization behavior~\cite{hanadjusting, wu2024freeinit}, we adopt a frequency decomposed initialization strategy that blends the low-frequency components of a reference image with high-frequency Gaussian noise. Specifically, given a reference image $\mathbf{X}_r$, we construct the initial latent noise $\mathbf{x}_T$ as:
\begin{equation}
    \mathbf{x}_T = \mathcal{F}^{-1}(\mathcal{M}_{\text{high}} \cdot \mathcal{F}(\epsilon) + \mathcal{M}_{\text{low}} \cdot \mathcal{F}(\mathbf{x}_r)),
    \label{eq:init}
\end{equation}
where $\mathbf{x}_r=E(\mathbf{X}_r)$ and $\mathcal{F}$ and $\mathcal{F}^{-1}$ denote the Fourier and inverse‐Fourier transforms, respectively, $\epsilon \sim \mathcal{N}(0, \mathbf{I})$ is Gaussian noise, and $\mathcal{M}_{\text{low}}, \mathcal{M}_{\text{high}}$ are complementary frequency masks (\textit{i.e.}, low‐pass and a high‐pass filter, respectively) that partition the frequency spectrum. In this initialization, the high-frequency components of $\mathbf{x}_T$ are drawn from the noise $\epsilon$, encouraging stochasticity. The low-frequency components from the reference image latent $\mathbf{x}_r$ introduce novel structural patterns. By decoupling broad spatial structure, which tends to be memorized, from semantics at the noise level, we reduce the model's reliance on memorized spatial priors during early denoising steps.

\mysection{Reference Image Selection.}
To obtain the reference image $\mathbf{X}_r$ used in initialization (\cref{eq:init}) and feature injection, CAPTAIN employs an online retrieval strategy to collect semantically related yet previously unseen images from the web. Given a prompt $\mathbf{p}$, we identify the top-$k$ visually attended words $\{w_i\}_{i=1}^k$ by aggregating the U-Net cross-attention maps over spatial queries, heads, layers, and timesteps, and pooling attention scores from tokens to words. In practice, we set $k=3$, and one word $w^{*}$ is randomly sampled from $\{w_i\}_{i=1}^k$ to serve as the retrieval query. Using $w^{*}$ as a keyword, we fetch a candidate set of web images $\mathcal{X}_{\text{web}} = \{\mathbf{X}_j\}$ from public APIs such as Pexels or Unsplash~\cite{Pexels, unsplash}. Each candidate and the query word are encoded using the CLIP vision ($V$) and text ($T$) encoders to obtain normalized embeddings $ \mathbf{f}_j = V(\mathbf{X}_j)$ and $\mathbf{g}^{*} = T(w^{*})$. Their cosine similarity defines the semantic alignment
$h_{1}(\mathbf{X}_j) = \cos(\mathbf{f}_j, \mathbf{g}^{*})$. To estimate dataset novelty, we precompute a FAISS index~\cite{johnson2019billion} of one million LAION-5B CLIP embeddings and define:
\begin{equation}
h_{2}(\mathbf{X}_j) =
1 - \max_{\ell \in\text{FAISS}} \cos(\mathbf{f}_j, \mathbf{f}_\ell).
\end{equation}
Additionally, we compute a perceptual uniqueness term $h_{3}$ from 64-bit perceptual hashes (pHash) as the normalized minimum Hamming distance between each candidate image and hashes from the LAION subset, rewarding images that are visually distinct from training samples. Each image receives a final composite score combining semantic relevance, dataset novelty, and perceptual uniqueness, and the final reference image is selected as:
\begin{equation}
\mathbf{X}_r =
\arg\max_{\mathbf{X}_j \in \mathcal{X}_{\text{web}}}
(\lambda_1 h_1 + \lambda_2 h_2 + \lambda_3 h_3),
\end{equation}
where $\boldsymbol{\lambda} = [0.3,\,0.4,\,0.3]$.
Alternatively, $\mathbf{X}_r$ can be directly provided by the user in our pipeline. For more details on our online image retrieval approach, refer to the Supplementary Material. Once $\mathbf{X}_r$ is obtained, we compute its latent representation $\mathbf{x}_r = E(\mathbf{X}_r)$ using the diffusion encoder, and use this semantic feature for injection during frequency decomposed noise initialization and 
the denoising process.

\begin{figure}[t]
\centering
\includegraphics[width=0.75\columnwidth]{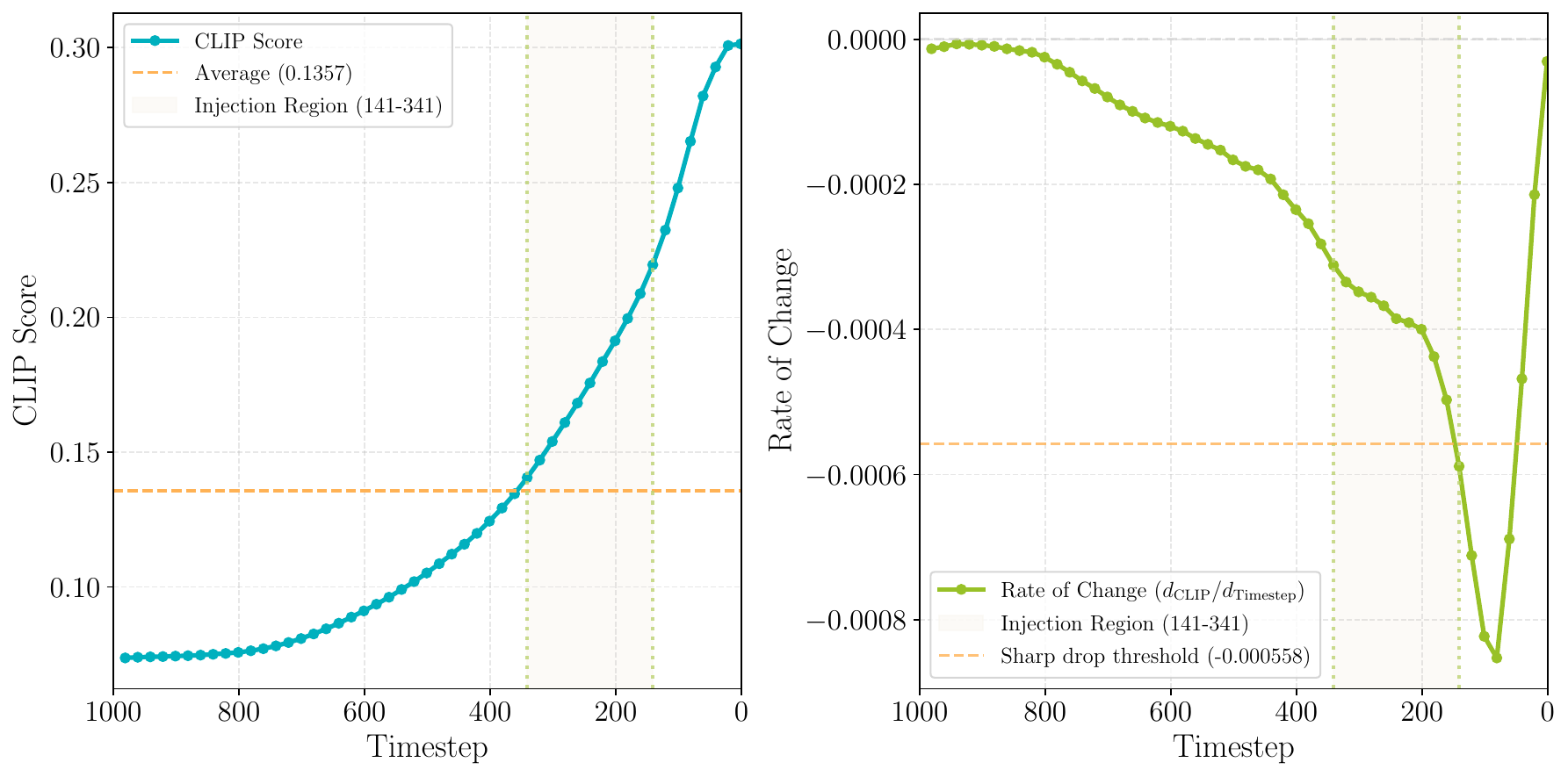}
    \caption{Identifying the optimal injection region for semantic editing. (Left) Average CLIP score evolution across diffusion timesteps for all the sdv1-500 memorized dataset. The injection region is bounded by the upper limit, where the CLIP score exceeds the average (indicating strong semantic content), and the lower limit before the sharp drop in the rate of change. (Right) First derivative of CLIP scores showing the rate of change. The sharp drop threshold (orange dashed line) identifies the transition point where fine details begin to stabilize, marking the end of the optimal injection window.}
    \label{fig:TML}
\end{figure}

\subsection{Timestep Window Injection Localization}
\label{sec:TML}

Given the latent features $\mathbf{x}_r$, the next step is to determine the optimal timesteps for feature injection during denoising. The diffusion process follows a hierarchical generation pattern: early timesteps establish coarse global structure and composition, while later ones progressively refine local details and textures \cite{ho2020denoising}. Memorization is most likely to occur during the last refinement phase, where the model may inadvertently reproduce specific visual cues from training samples once the overall structure has been formed.

To identify the optimal injection window, we analyze how image–text alignment changes over the denoising trajectory. At each timestep $t$, the partially denoised latent $\mathbf{x}_t$ is decoded to image space and compared with the conditioning prompt $\mathbf{p}$ using CLIP similarity:
\begin{equation}
    s_t = 
    \frac{
    V(D(\mathbf{x}_t))
    \cdot 
    T(\mathbf{p})
    }{
    \|V(D(\mathbf{x}_t))\|_2
    \|T(\mathbf{p})\|_2
    }.
\end{equation}
The similarity $s_t$ measures how well the semantic content of the prompt is expressed at timestep $t$. \Cref{fig:TML} (left) shows the average $s_t$ computed over the SDv1-500 memorized dataset. The injection window is bounded by the point at which the CLIP similarity first exceeds its mean (upper limit, $t_{\text{high}}$) and the point preceding the sharp decline in its rate of change (lower limit, $t_{\text{low}}$). The derivative of $s_t$ across timesteps (\cref{fig:TML} right) highlights this transition region, where a noticeable drop in $\mathrm{d}s_t / \mathrm{d}t$ indicates that fine-grained details start to converge. Then, the lower bound $t_{\text{low}}$ is determined as the first timestep where the derivative falls below a threshold defined as 
\begin{equation}
    t_{\text{low}} = \min_t \left\{\, \tfrac{\mathrm{d}s_t}{\mathrm{d}t} < \mu_{\mathrm{d}s_t/\mathrm{d}t} - 1.5\,\sigma_{\mathrm{d}s_t/\mathrm{d}t} \,\right\},
\end{equation}
where $\mu_{\mathrm{d}s_t/\mathrm{d}t}$ and $\sigma_{\mathrm{d}s_t/\mathrm{d}t}$ denote the mean and standard deviation of the derivative, respectively. Based on this observation, we define the timestep injection window as the range $t \in [t_{\text{low}}, t_{\text{high}}] = [141, 341]$, corresponding to the phase in which semantic alignment stabilizes while visual details are still being formed. Therefore, we inject features within this window to mitigate memorization. Note that this CLIP-based measure serves as a proxy for semantic consolidation rather than a direct indicator of memorization, and the exact timestep range may vary across diffusion architectures and datasets.

\subsection{Spatial Memorization Localization}
\label{sec:SML}

Once the timestep injection window $t \in [t_{\text{low}}, t_{\text{high}}]$ is determined, we aim to inject the reference features $\mathbf{x}_r$ into spatial regions that exhibit memorized content while remaining semantically consistent with $\mathbf{x}_r$. To this end, we employ two complementary mechanisms: the Bright Ending (BE) mask~\cite{chen2025exploring} and concept-specific attention maps~\cite{helbling2025conceptattention}. The BE mask $\mathbf{m}_{\text{BE}} \in [0,1]$ is obtained by extracting the cross-attention map of the final text token during a pseudo run at the last inference step of the diffusion model, where memorized image patches exhibit significantly greater attention to the final text token than non-memorized ones. 

\noindent To refine this localization with semantic awareness, we introduce a concept-specific attention extraction scheme adapted to the U-Net backbone. Given the target concept $w^{*}$ (obtained in \cref{sec:init}), we encode it with the diffusion model's text encoder and append its tokens to the standard prompt embeddings during cross-attention computation. We record cross-attention tensors from 16 layers across the U-Net's downsampling, middle, and upsampling blocks. We then average across attention heads, layers, and concept tokens, and normalize the aggregated map. After applying Gaussian smoothing with a standard deviation of 1.5, we obtain a concept mask $\mathbf{m}_{\text{concept}} \in [0,1]$. Finally, we compute the memorized concept region as the intersection between both masks:
\begin{equation}
\label{eq:mask_intersection}
\mathbf{m} = \mathbf{1}_{> \tau}\left( \mathbf{m}_{\text{BE}} \odot \mathbf{m}_{\text{concept}} \right), \ \mathbf{m} \in \left\{0,1\right\},
\end{equation}
where $\odot$ denotes element-wise multiplication, $\tau$ is a predefined threshold, and $\mathbf{1}_{>\tau}(\cdot)$ is the indicator function that outputs $1$ when its argument exceeds $\tau$ and $0$ otherwise.




\subsection{Feature Injection for Memorization Mitigation}
\label{sec:mitigation}

At each injection timestep $t \in [t_{\text{low}}, t_{\text{high}}]$, we reconstruct the predicted clean latent $\hat{\mathbf{x}}_0$ from the current noisy latent $\mathbf{x}_t$ using \cref{eq:predicted_x0}. The reference latent 
$\mathbf{x}_r$ is then injected into the spatially localized region defined by the mask $\mathbf{m}$, producing the modified latent:
\begin{equation}
    \hat{\mathbf{x}}'_0 = (1 - \delta\, \mathbf{m}) \odot \hat{\mathbf{x}}_0 + \delta\, \mathbf{m} \odot \mathbf{x}_r,
\end{equation}
where $\delta \in [0,1]$ controls the injection strength. The updated clean prediction $\hat{\mathbf{x}}'_0$ is then substituted into \cref{eq:next_step_latent} to compute the next latent state $\mathbf{x}_{t-1}$. This localized feature blending allows CAPTAIN to suppress memorized content while preserving semantic and structural coherence throughout the diffusion trajectory.

\section{Experiments}
\label{sec:experiments}

\mysection{Experimental Setup.} In line with prior work on memorization in diffusion models~\cite{somepalli2023understanding, wen2024detecting, ren2024unveiling, hintersdorf2024finding, chen2025exploring}, we evaluate our method on Stable Diffusion (SD) v1.4~\cite{rombach2022high}. We use 500 prompts extracted from the LAION dataset~\cite{schuhmann2022laion} that are known to trigger memorization in SD v1.4 to evaluate the effectiveness of memorization mitigation. These prompts, provided by \cite{webster2023reproducible}, are associated with cases where the model reproduces near-identical training images.

\mysection{Evaluation metrics.} We evaluate memorization mitigation using the Self-Supervised Copy Detection (SSCD) score~\cite{pizzi2022self}, which measures object-level similarity between a generated image and its nearest neighbor in the training set (lower is better $\downarrow$). Image-text alignment is assessed with the CLIP score~\cite{radford2021learning}, which quantifies the semantic consistency between each generated image and its corresponding text prompt (higher is better $\uparrow$).


\mysection{Baselines.} We compare CAPTAIN against three state-of-the-art inference-time memorization mitigation methods. BE~\cite{chen2025exploring} and PRSS~\cite{chen2025enhancing}, which use attention-based masking to achieve spatially-aware memorization suppression under both local and global memorization scenarios. We also include ~\cite{wen2024detecting}, which mitigates memorization by adjusting prompt embeddings based on conditional noise prediction, and \cite{hanadjusting}, which modifies the initial noise before denoising, providing a complementary initialization-based mitigation strategy. All baseline methods are applied at inference time and do not require access to the training data, making them suitable for fair comparison with our proposed approaches.


\mysection{Implementation details.} For each diffusion model, we generate one image per memorized prompt using seed 0 with identical inference configurations across all baselines and our proposed method. Specifically, for Stable Diffusion v1.4, we employ the DDIM~\cite{song2021denoising} sampler with 50 sampling steps and a CFG scale of 7.5. 
Additional implementation details are provided in the supplementary material.

\mysection{Compute Overhead} Our approach adds minimal overhead beyond standard Stable Diffusion inference. Running 500 prompts on a single A100 GPU took under 30 minutes, including noise initialization and mask computation. The full pipeline requires roughly 3 seconds per image, preserving practical inference speed. A runtime comparison with baseline methods is provided in the supplementary material.


\subsection{Comparison with baselines}
\begin{figure*}[t]
\centering
\begin{subfigure}[t]{0.32\textwidth}
    \centering
    \includegraphics[width=\linewidth]{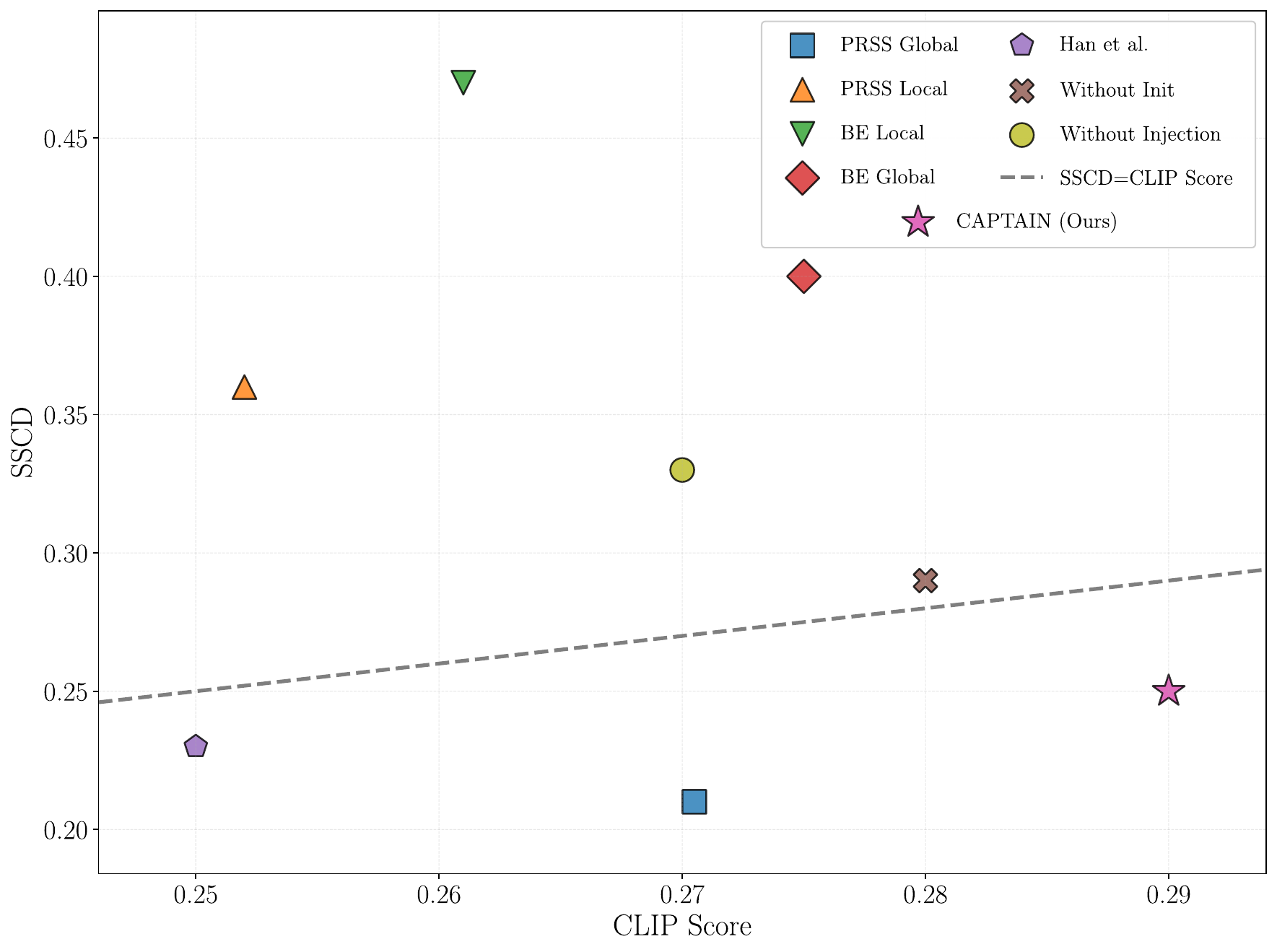}
    \caption{}
    \label{fig:main_quantitative}
\end{subfigure}\hfill
\begin{subfigure}[t]{0.32\textwidth}
    \centering
    \includegraphics[width=\linewidth]{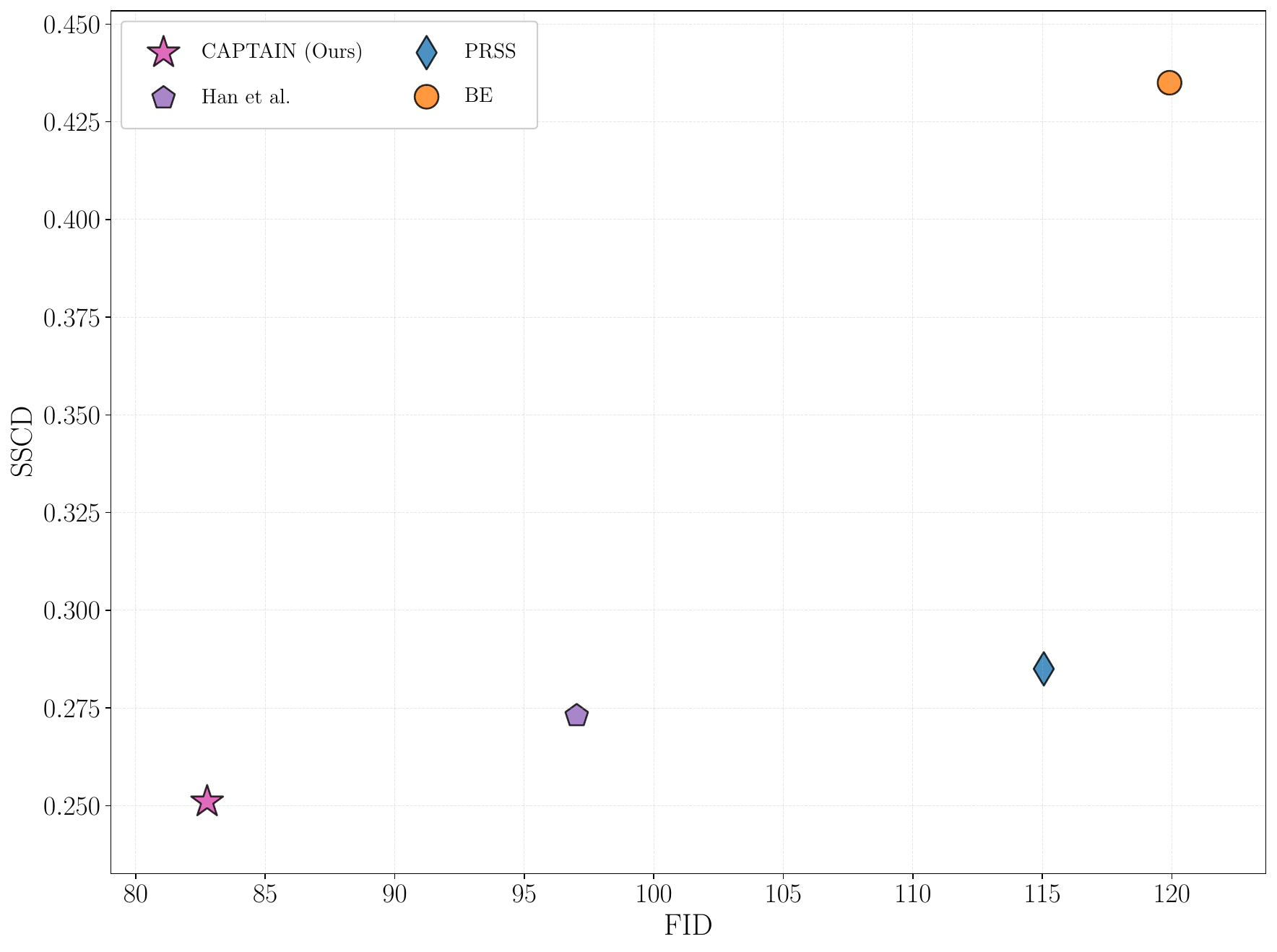}
    \caption{}
    \label{fig:FID}
\end{subfigure}\hfill
\begin{subfigure}[t]{0.32\textwidth}
    \centering
    \includegraphics[width=\linewidth]{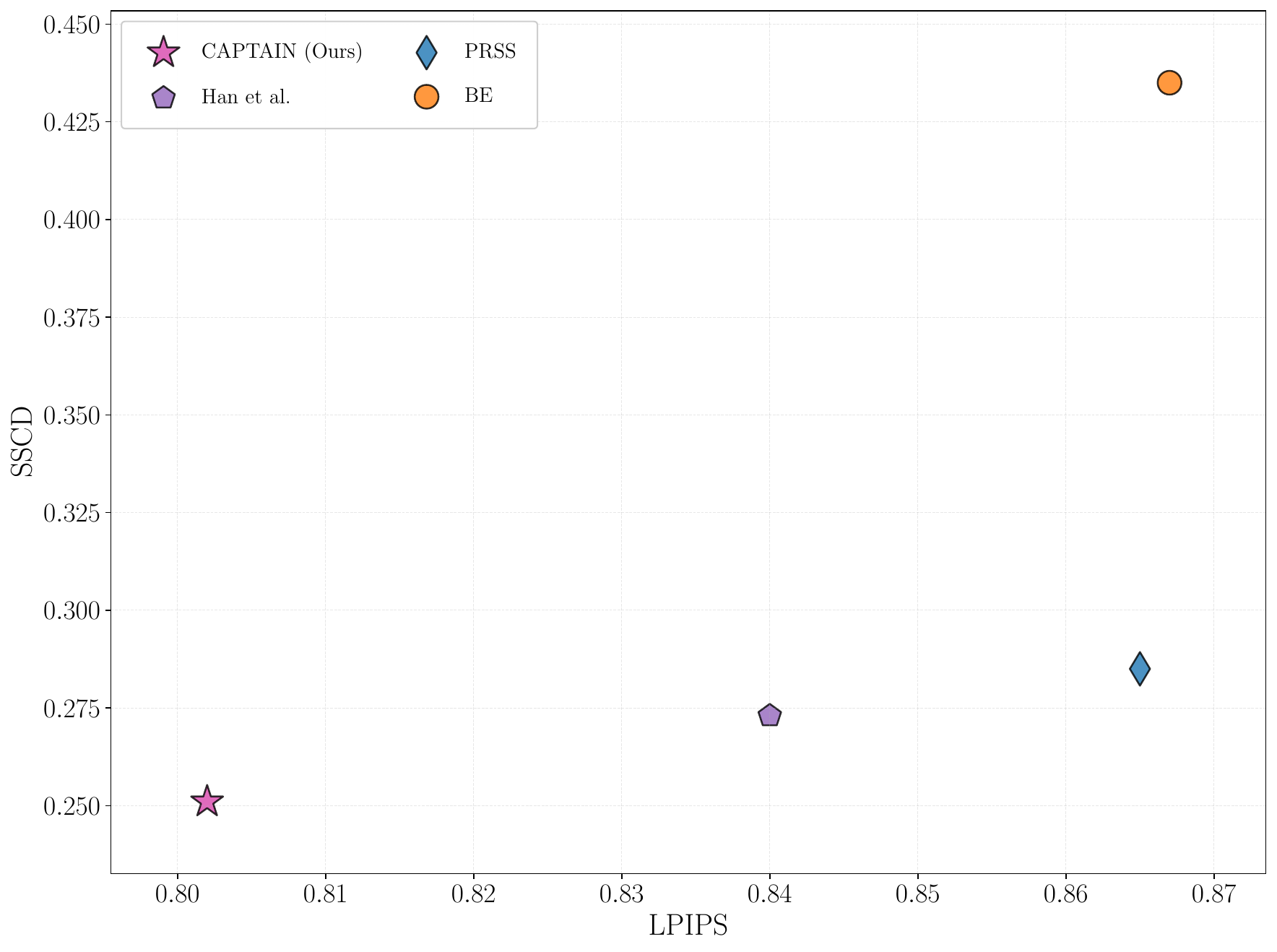}
    \caption{}
    \label{fig:LPIPS}
\end{subfigure}
\caption{Quantitative comparison of memorization mitigation methods under Stable Diffusion v1.4. (a) SSCD and CLIP scores across different methods. Lower SSCD indicates stronger memorization mitigation, while higher CLIP scores indicate better image–text alignment. (b) SSCD vs.\ FID, measuring image fidelity. (c) SSCD vs.\ LPIPS, measuring perceptual similarity. Lower values are preferred for all metrics except CLIP. Our method achieves the lowest SSCD while maintaining competitive fidelity and alignment, outperforming BE~\cite{chen2025exploring}, PRSS~\cite{chen2025enhancing}, , and Han \etal~\cite{hanadjusting}.}
\label{fig:combined_metrics}
\end{figure*}

\mysection{Quantitative Results.} \Cref{fig:main_quantitative} compares the performance of CAPTAIN against BE~\cite{chen2025exploring}, PRSS~\cite{chen2025enhancing}, and ~\cite{hanadjusting} using SSCD and CLIP scores. The diagonal dashed line ($\mathrm{SSCD} = \mathrm{CLIP}$ score) serves as a reference boundary, with points below this line indicating a favorable tradeoff between image–text alignment and memorization mitigation. As shown in \cref{fig:main_quantitative}, PRSS-Global and Han\etal achieve low SSCD scores (0.21–0.23), indicating strong memorization mitigation, but this comes at the cost of reduced semantic alignment, with CLIP scores of only 0.25–0.27. BE-Global attains the highest semantic alignment among baselines (CLIP $\approx 0.275$) but exhibits high SSCD values ($\approx 0.40$), demonstrating limited ability to suppress memorization. Finally, PRSS-Local and BE-Local occupy an intermediate region, with moderate CLIP (0.25–0.26) and higher SSCD (0.35–0.45), but neither metric shows favorable performance. In contrast, CAPTAIN achieves the strongest tradeoff, with the highest CLIP score (0.29) and the lowest SSCD among all non-overly-degrading methods (0.25). CAPTAIN improves semantic alignment by 28\% over PRSS-Global (0.29 vs. 0.21 CLIP) while maintaining comparable SSCD, and reduces SSCD by 38\% relative to BE-Global (0.25 vs. 0.40) without sacrificing prompt fidelity. Its position well below the $\mathrm{SSCD} = \mathrm{CLIP}$ reference line reflects effective decoupling of semantic alignment from structural memorization. These gains stem from combining noise initialization and localized feature injection. Unlike CFG-based methods (PRSS, BE) or initialization-only approaches (Han et al.), CAPTAIN directly guides the denoising trajectory away from memorized outputs by intervening at multiple stages of the generation process, enabling more effective, targeted suppression of memorization.

To further evaluate the image quality of our generated results and the baselines, we report FID and LPIPS scores. As shown in~\cref{fig:FID} and ~\cref{fig:LPIPS}, our method achieves the lowest values across both metrics, indicating higher visual quality while still mitigating memorization. For PRSS and BE, we take their global- and local-variant SSCD scores from Fig.~\ref{fig:main_quantitative}, average them, and use the averages as their SSCD points in this comparison.

\begin{figure*}[t]
\centering
\includegraphics[width=\linewidth]{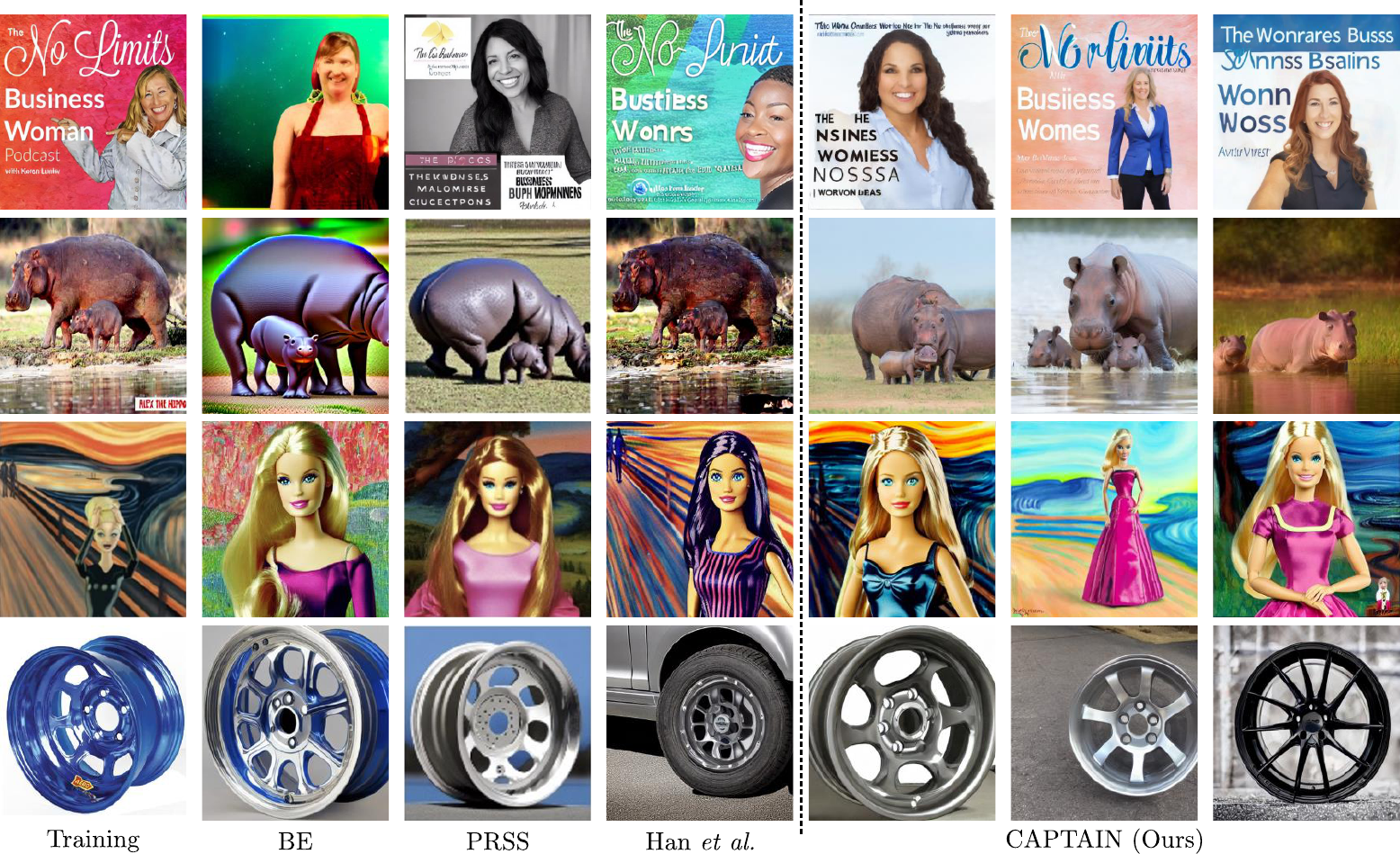}
    \caption{Qualitative comparison with different memorization mitigation methods: 
    BE(\cite{chen2025exploring}), 
    PRSS(\cite{chen2025enhancing}), and Han et al.~\cite{hanadjusting}. The last three columns show results from our approach using different random seeds. The prompts used for generation and more qualitative results are provided in the supplementary material.}
    \label{fig:main_qualitative}
\end{figure*}

\mysection{Qualitative Results.} \Cref{fig:main_qualitative} compares our approach with baseline methods, including BE(\cite{chen2025exploring}), PRSS(\cite{chen2025enhancing} and ~\cite{hanadjusting}. Each row illustrates a representative memorization case from the SDv1-500 benchmark \cite{webster2023reproducible}. Across diverse scenarios—including portraits, animals, artistic compositions, and product imagery—baseline methods reduce some degree of duplication but often preserve key memorized structures or introduce noticeable degradations. As observed, our approach effectively suppresses memorized content while producing visually coherent and prompt-faithful results across different random seeds (rightmost columns). Examples include removing the distinctive facial attributes of training portraits, altering animal poses and appearances, and generating novel object shapes while preserving category-level semantics. Additional qualitative examples and prompt lists are provided in the supplementary material.

\begin{figure*}[t]
    \centering
    \includegraphics[width=\linewidth]{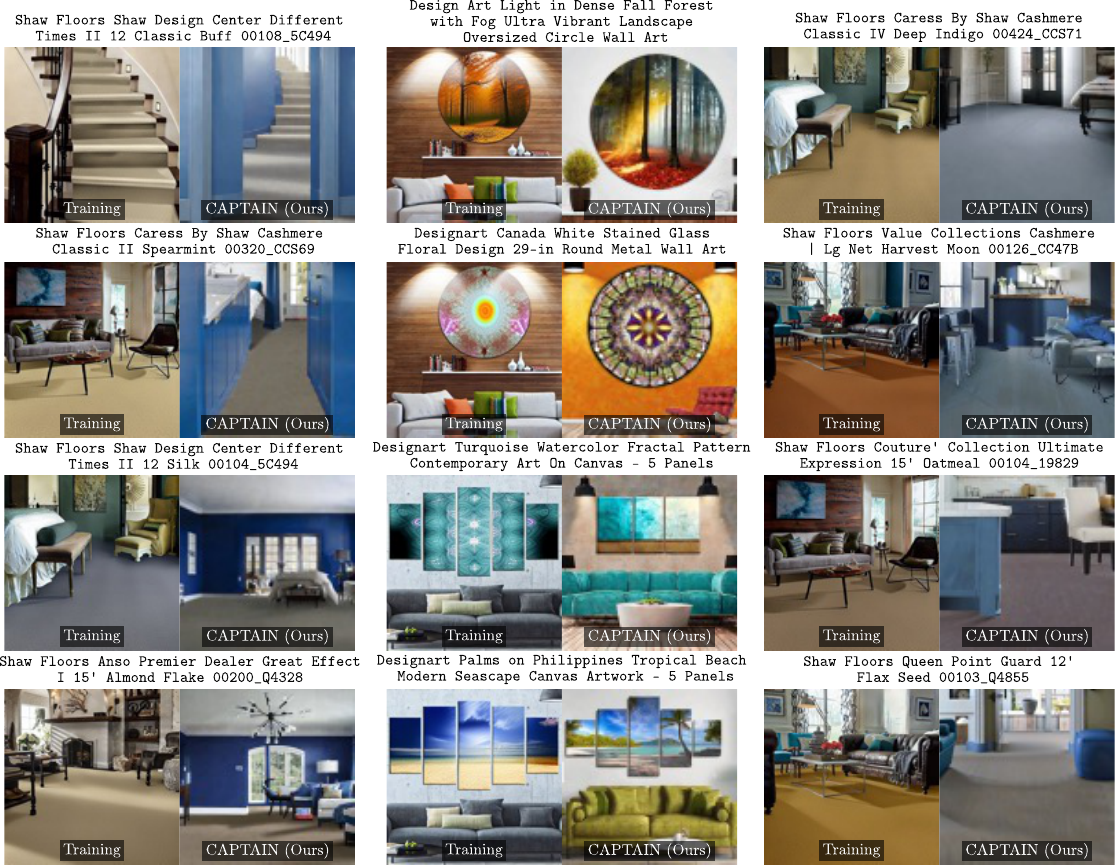}
    \caption{CAPTAIN also shows strong semantic alignment while reducing memorization on Stable Diffusion 2.0 (SD~2.0). In each pair, the left image shows the memorized output produced by SD~2.0, and the right image is the mitigated result produced by our approach. The prompt used for generation is shown above each pair.}
    \label{fig:sd2}
\end{figure*}

\mysection{Evaluation on Stable Diffusion 2.0.}
We further evaluate our approach on Stable Diffusion 2.0\footnote{\url{https://huggingface.co/lzyvegetable/stable-diffusion-2-1}}. Since SD~2.0 is trained on a de-duplicated dataset, our method achieves even better SSCD performance (0.202) on the 219 non-memorized samples of \cite{ren2024unveiling}, while maintaining strong semantic alignment (CLIP score 0.258). Compared to the results reported in ~\cite{hanadjusting} (0.248 CLIP, 0.226 SSCD), this corresponds to a \textbf{4.03\% increase} in CLIP and a \textbf{10.84\% decrease} in SSCD. Visual comparisons between CAPTAIN and the memorized outputs are shown in Fig.~\ref{fig:sd2}.

\begin{table}[t]
\centering
\small
\begin{subtable}[t]{0.6\columnwidth}
  \centering
  \caption{Init. and feat. inj. ablation.}
  \label{tab:ablations}
  \begin{tabular}{cc|cc|cc}
    \toprule
          &            & \multicolumn{2}{c}{$\delta = 0.1$} & \multicolumn{2}{c}{$\delta = 0.2$} \\ 
    \midrule
    Init. & Feat. Inj. & CLIP $\uparrow$ & SSCD $\downarrow$ & CLIP $\uparrow$ & SSCD $\downarrow$ \\ 
    \midrule
    \cmark   & \xmark    & 0.279 & 0.291 & 0.279 & 0.291 \\
    \xmark   & \cmark    & 0.270 & 0.332 & 0.232 & 0.247 \\
    \rowcolor{green!10} \cmark & \cmark & 0.292 & 0.250 & 0.262 & 0.222 \\
    \bottomrule
  \end{tabular}%
\end{subtable}
\hfill
\begin{subtable}[t]{0.35\columnwidth}
  \centering
  \caption{Mask ratio ablation.}
  \label{tab:mask_ablation}
  \begin{tabular}{@{}lcc@{}}
    \toprule
    $\tau$ & CLIP $\uparrow$ & SSCD $\downarrow$ \\
    \midrule
    0.1 & 0.292 & 0.250 \\
    0.2 & 0.275 & 0.256 \\
    0.3 & 0.253 & 0.251 \\
    0.4 & 0.250 & 0.243 \\
    0.5 & 0.244 & 0.236 \\
    \bottomrule
  \end{tabular}
\end{subtable}
\vspace{0.5cm}
\caption{Ablation studies on our method. (a) Impact of initialization and feature injection across different $\delta$ values. Our full approach (highlighted) is shown in the last row. (b) Effect of different mask ratios $\tau$ on performance.}
\label{tab:combined_ablations}
\end{table}

\subsection{Ablation Studies}

\Cref{tab:ablations} evaluates the contribution of CAPTAIN's two core components, noise initialization and feature injection, using SSCD and CLIP scores under different injection strengths $\delta$. \Cref{fig:main_quantitative} also shows the ablation results when using our method without Noise initialization (Without Init) and without feature injection (Without Injection).

\mysection{Noise Initialization.} When feature injection is disabled, the perturbation strength $\delta$ has no effect on the generation process, since $\delta$ only scales the injected features. As a result, the initialization-only row in \cref{tab:ablations} yields identical CLIP and SSCD scores across all $\delta$ values. This shows that initialization provides a fixed level of memorization mitigation (SSCD $\approx 0.29$) but cannot be strengthened or adapted during denoising. In practice, initialization establishes a static mitigation effect at the start of the diffusion trajectory, yet lacks the flexibility required to further reduce memorization in later steps.


\mysection{Feature Injection.} Feature injection acts as the dynamic, adaptive component of CAPTAIN, intervening at intermediate stages of the diffusion process to modify latent features. Without initialization, however, injection becomes highly sensitive to the perturbation strength $\delta$. At low $\delta$, it preserves semantic alignment (high CLIP) but fails to suppress memorization (SSCD = 0.332). At high $\delta$, the opposite occurs, memorization decreases (SSCD = 0.179) but semantic alignment degrades sharply (CLIP = 0.201). This instability indicates that feature injection alone cannot reliably balance alignment and memorization mitigation across different perturbation levels.

\mysection{Combined Effect.} The full CAPTAIN configuration (initialization + injection) achieves the best trade-off across all $\delta$ values, consistently improving privacy while preserving semantic alignment (e.g., CLIP = 0.292, SSCD = 0.250 at $\delta=0.1$). Initialization stabilizes the injection mechanism, while injection provides the adaptability that initialization alone lacks. Together, they yield predictable, well-balanced behavior that neither component achieves independently.

\mysection{Mask Threshold. }We also conduct an ablation study on the $\tau$ threshold defined in \cref{eq:mask_intersection}. In practice, we set $\tau = 0.1$, meaning that only values greater than 0.1 after the element-wise multiplication are retained in the binary mask. To ensure a valid mask during injection, we handle the rare case where thresholding produces an all-zero mask, which occurs when every element of the product falls below $\tau$. In such cases, we fall back to the BE mask and apply a threshold of 0.5 to obtain a valid binary mask. We additionally evaluate $\tau = 0.2, 0.3, 0.4, 0.5$. Table~\ref{tab:mask_ablation} shows that increasing $\tau$ results in lower SSCD, but also leads to a noticeable drop in the CLIP score compared to the default $\tau=0.1$ setting. This reflects weaker semantic alignment between the generated image and the prompt. Considering this trade-off, we adopt $\tau=0.1$ as the most stable and balanced value.
\section{Conclusions}
\label{sec:conclusions}

We introduced CAPTAIN, a training-free framework for mitigating memorization in text-to-image diffusion models. Unlike prior inference-time methods that rely on manipulating CFG strength or perturbing prompt embeddings, CAPTAIN operates directly in the latent space. Our frequency-based noise initialization reduces the tendency to reproduce memorized patterns in early denoising steps, while our temporal–spatial localization strategy identifies effective timesteps and regions for intervention. By injecting semantically aligned features from non-memorized reference images, CAPTAIN suppresses memorized content without sacrificing visual quality or alignment with the conditioning prompt. Extensive experiments demonstrate that CAPTAIN achieves stronger memorization mitigation than existing inference-time baselines while maintaining competitive semantic fidelity.

\mysection{Limitations.} We acknowledge our approach presents some limitations. First, it relies on retrieving an external reference image, and variations in retrieval quality may influence injection effectiveness. Second, our spatial localization strategy, based on BE attention and concept-specific attention, can be less reliable for abstract or ambiguous prompts, occasionally producing overly small or diffuse masks. Third, CAPTAIN requires a FAISS index to estimate the novelty of a retrieved image; while we will provide one for Stable Diffusion, a new index must be built when applying the method to other models trained on different datasets. Finally, although computationally lightweight, CAPTAIN introduces extra operations such as frequency decomposition and CLIP-based timestep selection.

\mysection{Ethical and Legal Considerations.} CAPTAIN retrieves reference images from public, royalty-free APIs (Pexels and Unsplash), which provide openly licensed content for research and creative use. No personal, private, or copyrighted datasets are accessed or stored. Retrieved images are used exclusively at inference time to guide feature injection and are not included in training or redistributed. This design aligns with CAPTAIN's goal of reducing copyright and privacy risks associated with memorization in diffusion models.

\mysection{Acknowledgments}. The research reported in this publication was supported by funding from King Abdullah University of Science and Technology (KAUST) - Center of Excellence for Generative AI, under award number 5940.

{
    \small
    \bibliographystyle{unsrtnat}
    \bibliography{main}
}

\appendix

\clearpage
\setcounter{page}{1}



\section{Reference Image Selection Details}

This section details our reference-image retrieval procedure, which identifies a concept word from cross-attention, retrieves semantically related web images, and selects a novel, perceptually distinct reference. The full procedure is summarized in Algorithm~\ref{algo:ref_image_selection}.

\subsection{Attention-Based Query Word Extraction.}
Given a prompt $\mathbf{p}$, CAPTAIN identifies visually relevant words by aggregating U-Net cross-attention across layers, timesteps, attention heads, and spatial queries. For each layer $l$, timestep $t$, and head index $u$, the cross-attention matrix
\(A^{(l,t,u)} \in \mathbb{R}^{Q \times N_{\text{tok}}}\)
captures attention from the $N_{\text{tok}}$ text tokens to the $Q$ spatial query positions. We compute:
\begin{equation}
\mathbf{s}_{\text{tok}} = \sum_{l,t} \frac{1}{H\,Q} \sum_{u=1}^{H} \sum_{q=1}^{Q} A^{(l,t,u)}_{q,:},
\label{eq:crossattn}
\end{equation}
yielding token-level importance scores. To obtain word-level scores, token scores are mapped to their corresponding words in the prompt (preserving order), producing the unfiltered word-score
vector $\tilde{\mathbf{s}}_{\text{word}}$. A filtering operator
$\Phi(\cdot)$ removes stopwords and punctuation:
\begin{equation}
\mathbf{s}_{\text{word}} = \Phi(\tilde{\mathbf{s}}_{\text{word}}).
\end{equation}
Finally, the top-$k$ highest-scoring words are selected. In practice, one word $w^{*}$ (with $k=3$) is randomly sampled to form the retrieval query.

\subsection{Online Image Retrieval and Scoring.}
Using $w^{*}$ as a keyword, we retrieve a candidate set $\mathcal{X}_{\text{web}}=\{\mathbf{X}_j\}$ from public APIs (e.g., Pexels, Unsplash). To further reduce the likelihood that retrieved samples overlap with the Stable Diffusion training data (drawn mainly from images up to 2022), we preferentially select images uploaded after 2024. Each candidate and the query word are encoded using CLIP vision ($V$) and text ($T$) encoders:
\begin{equation}
\mathbf{f}_j = V(\mathbf{X}_j), \qquad
\mathbf{g}^{*} = T(w^{*}),
\label{eq:clip_encode}
\end{equation}
and the semantic alignment score is defined as
\begin{equation}
h_1(\mathbf{X}_j) = \cos(\mathbf{f}_j, \mathbf{g}^{*}).
\label{eq:h1}
\end{equation}

\mysection{Dataset Novelty.}
To estimate novelty with respect to the data distribution seen by Stable Diffusion, we build a FAISS\cite{johnson2019billion} index using 1M CLIP ViT-L/14 embeddings randomly sampled from the HuggingFace LAION ``laion2B-en-aesthetic'' subset\footnote{\url{https://huggingface.co/datasets/laion/laion2B-en-aesthetic}}. This subset is derived from the larger LAION-5B corpus but represents only a filtered portion of the original dataset. We use a flat L2 index over unit-normalized embeddings and compute the novelty score as:
\begin{equation}
h_2(\mathbf{X}_j) = 1 - \max_{\ell \in \text{FAISS}}
\cos(\mathbf{f}_j, \mathbf{f}_{\ell}).
\label{eq:h2}
\end{equation}

\begin{algorithm*}[t]
\footnotesize
\SetAlgoLined
    \PyCode{\textcolor{magenta}{def} select\_reference\_image(prompt, lambda1, lambda2, lambda3):}\\
    \Indp
        \PyComment{1) Extract query word \cref{eq:crossattn}}\\
        \PyCode{attn = collect\_cross\_attention(prompt)}\\
        \PyCode{s\_tok = sum(A.mean(dim=0) \textcolor{magenta}{for} A \textcolor{magenta}{in} attn.values())}\\
        \PyCode{s\_word = tokens\_to\_words(s\_tok, prompt)}\\
        \PyCode{top3 = s\_word.topk(3).indices}\\
        \PyCode{w\_star = random.choice(top3)}\\

        \BlankLine
        \PyComment{2) Retrieve post-2024 web images}\\
        \PyCode{sources = ["Unsplash", "Pexels", }\\
        \Indp 
            \PyCode{"Wikimedia Commons"]} \\
        \Indm
        \PyCode{X\_web = query\_image\_apis(w\_star, }\\
        \Indp
            \PyCode{sources, min\_year=2024)} \\
        \Indm
        \PyCode{g\_star = encode\_text(w\_star)} \PyComment{\cref{eq:clip_encode}}\\

        \BlankLine
        \PyComment{3) Score candidates using (h1, h2, h3)}\\
        \PyCode{best\_score, X\_r = $-\infty$, \textcolor{magenta}{None}}\\
        \PyCode{\textcolor{magenta}{for} X\_j \textcolor{magenta}{in} X\_web:}\\
        \Indp
            \PyCode{f\_j = encode\_image(X\_j)} \PyComment{\cref{eq:clip_encode}}\\
            \PyCode{h1 = cos\_sim(f\_j, g\_star)} \PyComment{\cref{eq:h1}}\\
            \PyCode{h2 = 1 - faiss\_max\_sim(f\_j)} \PyComment{\cref{eq:h2}}\\
            \PyCode{pi\_j = phash64(X\_j)}\\
            \PyCode{d\_min = min\_hamming(pi\_j, laion\_phash)}\\
            \PyCode{h3 = clip(d\_min / 32)}\PyComment{\cref{eq:h3}}\\
            \PyCode{score = lambda1*h1 + lambda2*h2 + lambda3*h3}\\
            \PyCode{\textcolor{magenta}{if} score > best\_score:} \\
            \Indp
                \PyCode{best\_score, X\_r = score, X\_j}\\
            \Indm
        \Indm

        \BlankLine
        \PyCode{\textcolor{magenta}{return} X\_r}\\
    \Indm

\caption{Reference Image Selection.}
\label{algo:ref_image_selection}
\end{algorithm*}

\mysection{Perceptual Uniqueness.} We compute a 64-bit perceptual hash (pHash) for each candidate image $\mathbf{X}_j$, producing a binary hash $\pi_j$. To measure perceptual dissimilarity with respect to the model's training distribution, we precompute a large set of 64-bit pHashes from the laion2B-en-aesthetic subset and denote them by $\{\pi_{\ell}\}$. The perceptual uniqueness score is defined as the minimum normalized Hamming distance between $\pi_j$ and these LAION hashes:
\begin{equation}
h_3(\mathbf{X}_j) = \min_{\ell} \;\mathrm{clip}\!\left( \frac{\mathrm{Hamming}(\pi_j, \pi_{\ell})}{32},\,0,1 \right),
\label{eq:h3}
\end{equation}
where $\mathrm{clip}(\cdot,0,1)=\min(\max(\cdot,0),1)$ limits the normalized Hamming distance to $[0,1]$, and dividing by 32 corresponds to half the maximum 64-bit distance, leading larger distances to saturate at one after clipping. Larger values correspond to candidates that are more perceptually distinct from LAION samples, reducing the likelihood of overlap with training-set images.

\mysection{Final Selection.} Each candidate image receives a composite score formed by a weighted combination of semantic relevance, dataset novelty, and perceptual uniqueness:
\begin{equation}
\mathbf{X}_r = \arg\max_{\mathbf{X}_j\in\mathcal{X}_{\text{web}}}
\left( \lambda_1 h_1(\mathbf{X}_j) + \lambda_2 h_2(\mathbf{X}_j) + \lambda_3 h_3(\mathbf{X}_j) \right),
\label{eq:final_score}
\end{equation}
where $\boldsymbol{\lambda} = [0.3,\,0.4,\,0.3]$. Users may alternatively provide $\mathbf{X}_r$ directly. The latent reference representation $\mathbf{x}_r = E(\mathbf{X}_r)$ is then used for both noise initialization and feature injection in CAPTAIN.

\section{Computation Overhead Comparison}

We evaluate the computational overhead of our method against baseline models in Table~\ref{tab:overhead}, including BE~\cite{chen2025exploring}, PRSS~\cite{chen2025enhancing}, and ~\cite{hanadjusting}. All methods are compared against the vanilla Stable Diffusion v1.4, whose average inference time \textbf{2.06} is used to compute the per-method overhead. As reported in Section~\ref{sec:experiments}, our approach maintains computational efficiency in terms of both wall-clock time and FLOPs. BE and PRSS exhibit similar efficiency with minimal overhead. However, the per-sample mitigation variant of ~\cite{hanadjusting} exhibits significantly higher overhead than other methods due to its optimization-based noise initialization procedure, which incurs additional computational costs that vary across cases. Notably, their batch-wise variant achieves the fastest inference time among all methods, although it does not perform better than the per-sample variant as shown in \cite{hanadjusting}. Furthermore, as noted in \cite{hanadjusting}, per-sample initialization can be retrieved from cached previous initializations when a sample has been processed before, thereby reducing computational overhead in practice. Unless otherwise noted, in our experiments we mainly compare against the per-sample variant of ~\cite{hanadjusting}, as it outperforms the batch-wise variant.

\begin{table}[h]
\centering
\small
\caption{Computational overhead comparison over 10 runs with 50 inference steps on SD v1.4.}
\label{tab:overhead}
\begin{tabular}{lccc}
\toprule
\textbf{Method} & \textbf{GFLOPS} & \textbf{Time (s)} & \textbf{Overhead} \\
\midrule
~\cite{chen2025exploring}(BE) & 303.45 & $2.23 \pm 0.32$ & 1.08$\times$ \\
~\cite{chen2025enhancing}(PRSS) & 136.64 & $4.96\pm 0.48 $ & $2.41\times$ \\
 ~\cite{hanadjusting} [batch-wise]& 317.25 & $2.14 \pm 0.32 $ & $1.04\times$ \\
~\cite{hanadjusting} [per-sample] & 62.28 & $10.88 \pm 10.24$ & $5.28\times $\\
\midrule
\textbf{CAPTAIN (Ours)} & 235.83 & $2.87 \pm 0.87$ & $1.39\times$ \\
\bottomrule
\end{tabular}
\end{table}


\section{Additional Quantitative Result}



\subsection{Per-Sample Optimal Hyperparameter Results}

To evaluate the strongest performance our method can achieve, we run a full hyperparameter search for each prompt. For every example, we select the conditioning concept extracted from the prompt, choose the corresponding conditioning image, and optimize the injection strength $\delta$. Across 500 prompts, the best configuration per sample reaches a CLIP score of $0.294$ and an SSCD score of $0.213$. In comparison, our default setting (using $\delta=0.1$) achieves CLIP $0.292$ and SSCD $0.250$, as reported in the main paper. This corresponds to a small CLIP improvement of about 0.7\% and a larger 14.8\% reduction in SSCD, indicating lower similarity to training images while still preserving strong semantic alignment.




\begin{figure*}[t]
    \centering
    \includegraphics[width=\linewidth]{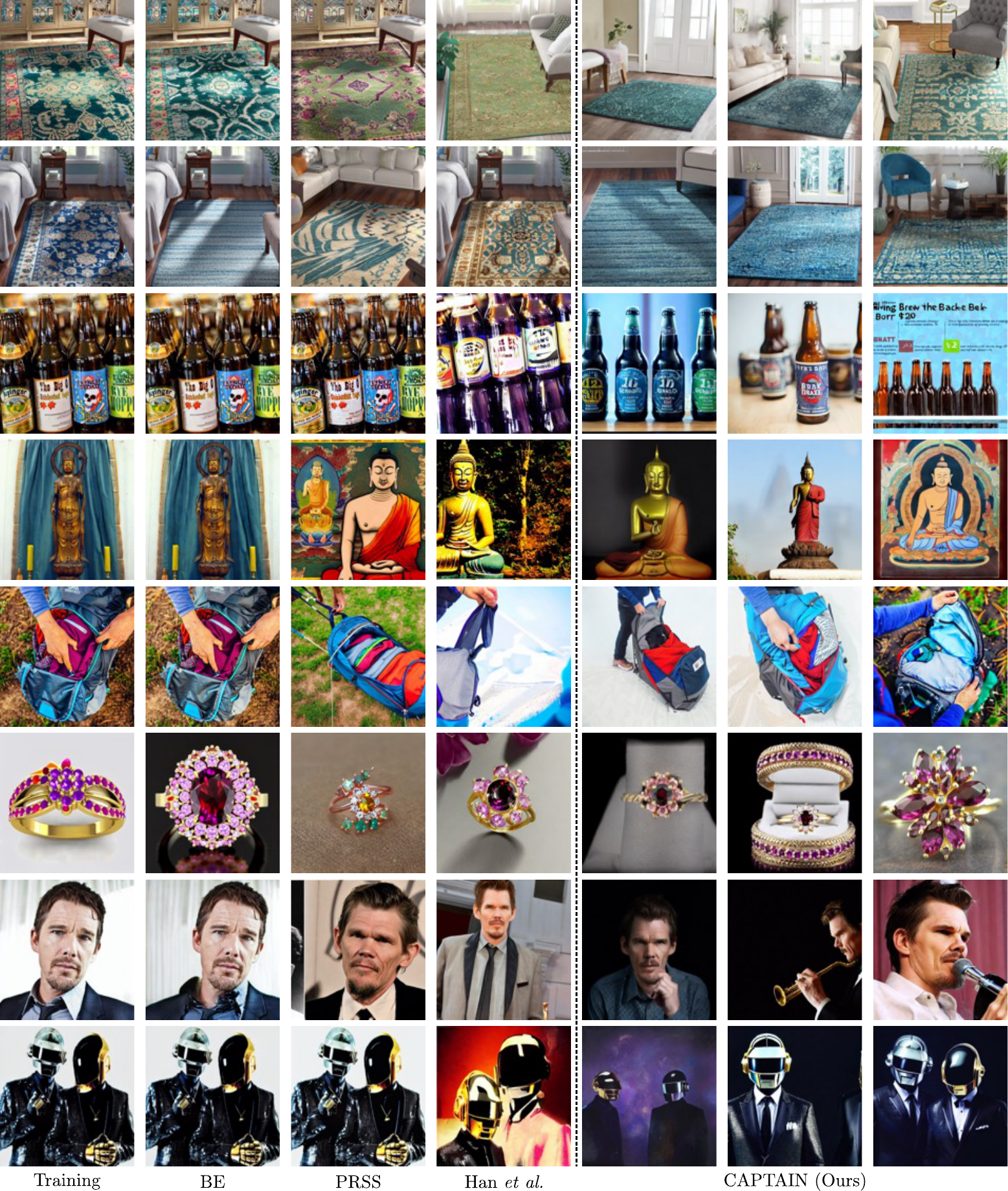}
    \caption{Additional Qualitative comparisons between CAPTAIN and the baseline methods, BE \cite{chen2025exploring}, PRSS\cite{chen2025enhancing}, and Han \etal \cite{hanadjusting}.}
    \label{fig:supp_qualitative1}
\end{figure*}
\begin{figure*}[t]
    \centering
    \includegraphics[width=\linewidth]{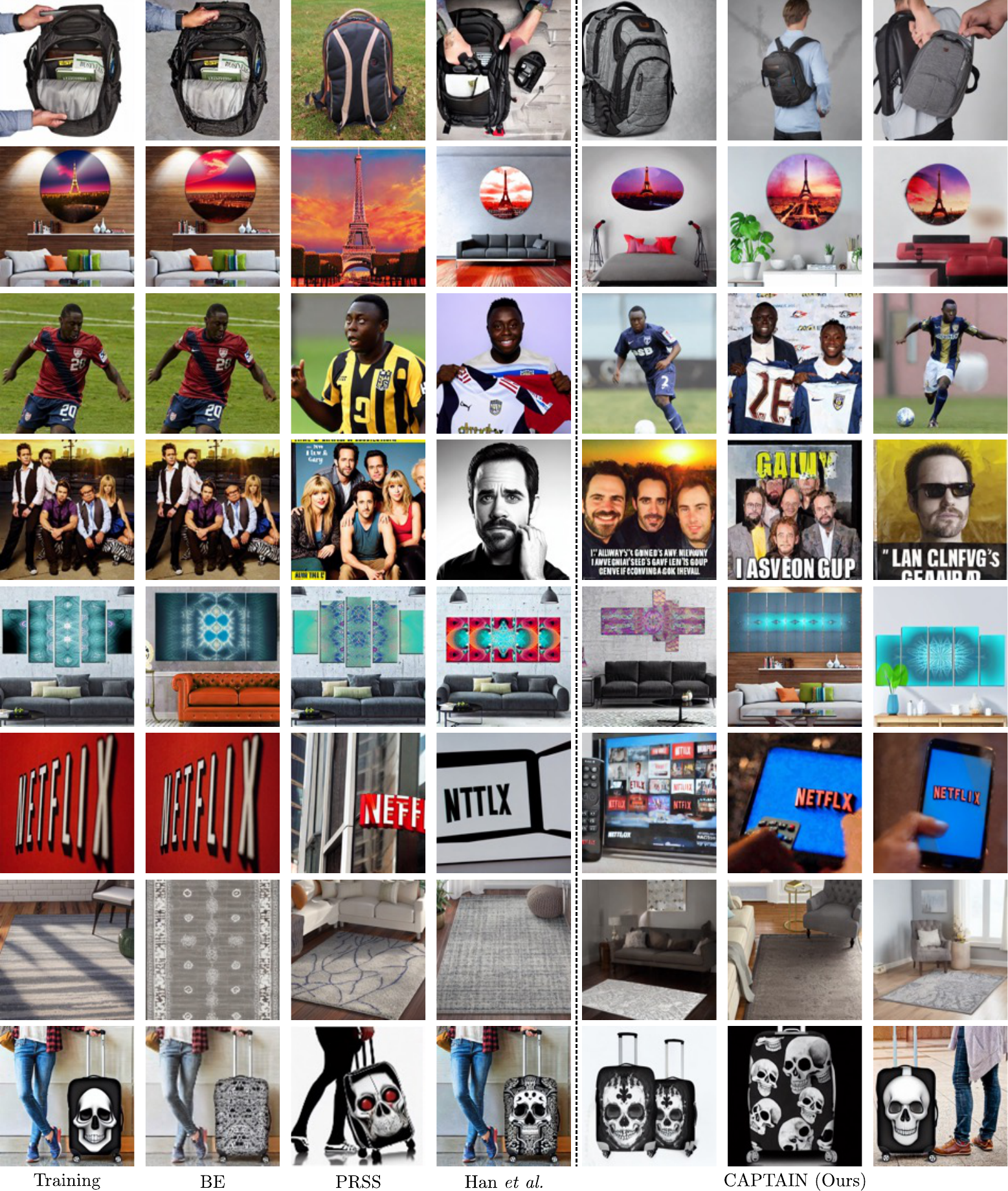}
    \caption{Additional Qualitative comparisons between CAPTAIN and the baseline methods, BE \cite{chen2025exploring}, PRSS\cite{chen2025enhancing}, and Han \etal \cite{hanadjusting}.}
    \label{fig:supp_qualitative2}
\end{figure*}

\begin{table*}[t]
\centering
\caption{Prompts associated with the visual examples in \cref{fig:main_qualitative}, \cref{fig:supp_qualitative1}, and \cref{fig:supp_qualitative2}. }
\label{tab:prompts}
\begin{subtable}[t]{0.48\textwidth}
\centering
\resizebox{0.98\linewidth}{!}{%
\begin{tabular}{C{3cm} C{7cm}}
\toprule
\textbf{Training Image} & \textbf{Prompt} \\
\midrule

\includegraphics[width=2.5cm]{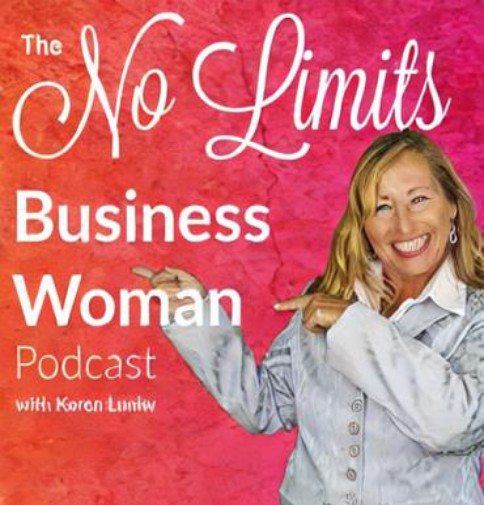} &
\texttt{The No Limits Business Woman Podcast} \\
\midrule
\includegraphics[width=2.5cm]{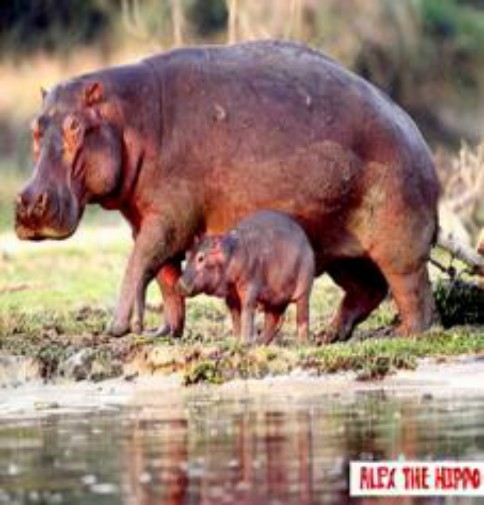} &
\texttt{Mothers influence on her young hippo} \\
\midrule

\includegraphics[width=2.5cm]{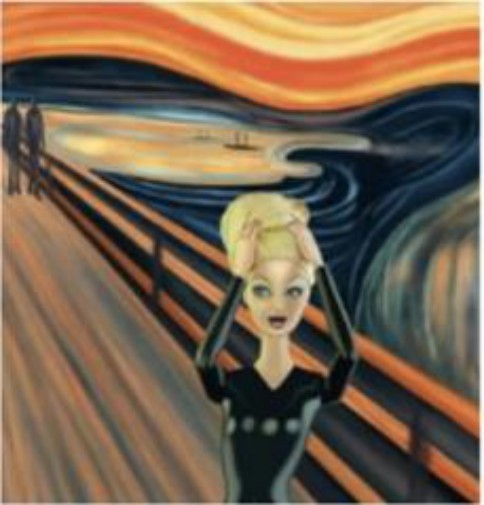} &
\texttt{If Barbie Were The Face of The World's Most Famous Paintings} \\
\midrule

\includegraphics[width=2.5cm]{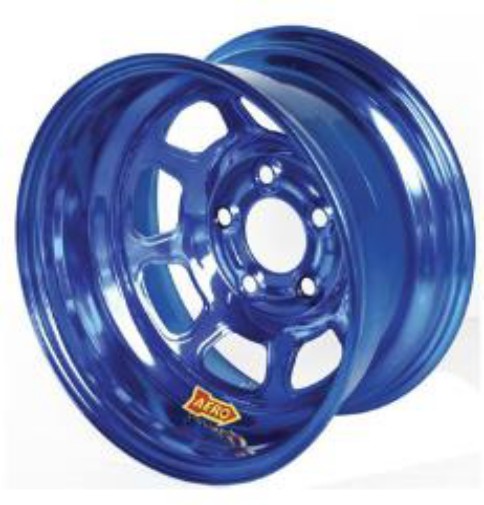} &
\texttt{Aero 52-984720BLU 52 Series 15×8 Wheel, 5 on 4–3/4 BP, 2 Inch BS IMCA} \\
\midrule

\includegraphics[width=2.5cm]{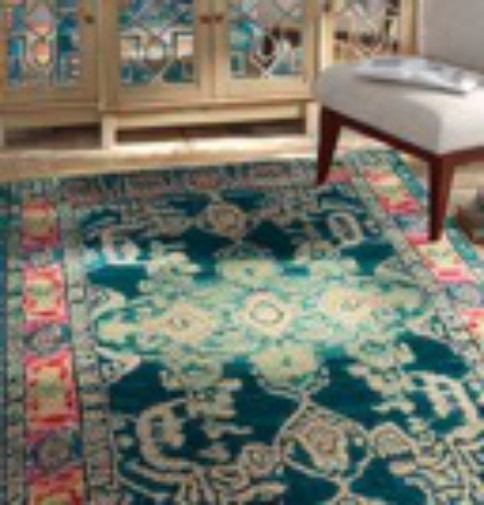} &
\texttt{Annabel Green Area Rug by Bungalow Rose} \\
\midrule

\includegraphics[width=2.5cm]{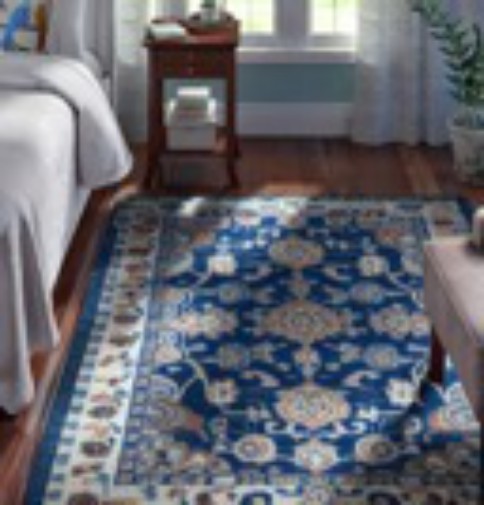} &
\texttt{Lilah Teal Blue Area Rug by Andover Mills} \\
\midrule

\includegraphics[width=2.5cm]{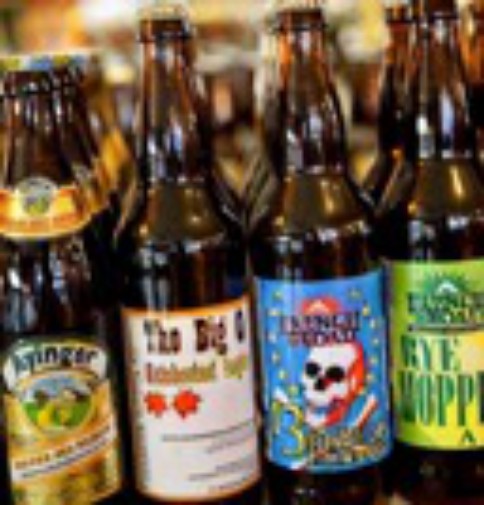} &
\texttt{Breaking Down the \$12 in Your Six–Pack of Craft Beer} \\
\midrule

\includegraphics[width=2.5cm]{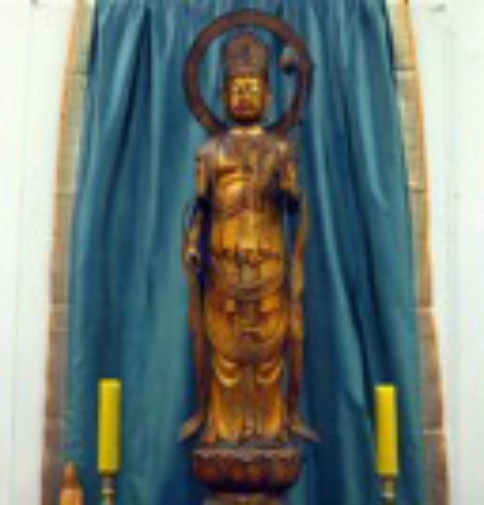} &
\texttt{Talks on the Precepts and Buddhist Ethics} \\
\midrule

\includegraphics[width=2.5cm]{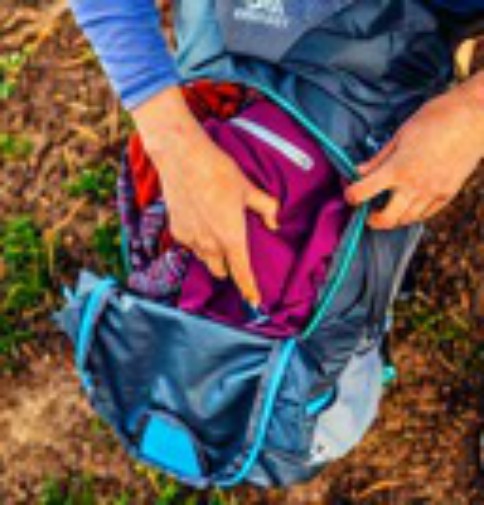} &
\texttt{Full body U-Zip main opening on front of bag for easy unloading when you get to camp} \\
\midrule

\includegraphics[width=2.5cm]{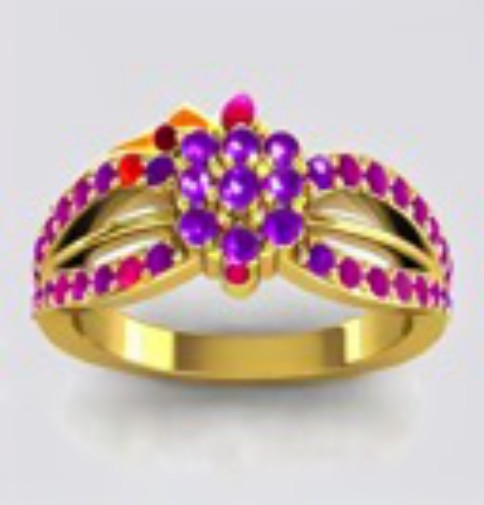} &
\texttt{Simple Floral Pave Utpala Garnet Ring with Amethyst and Pink Tourmaline in 18k Yellow Gold} \\
\bottomrule

\end{tabular}}
\end{subtable}
\hfill
\begin{subtable}[t]{0.48\textwidth}
\centering
\resizebox{0.98\linewidth}{!}{%
\begin{tabular}{C{3cm} C{7cm}}
\toprule
\textbf{Training Image} & \textbf{Prompt} \\
\midrule

\includegraphics[width=2.5cm]{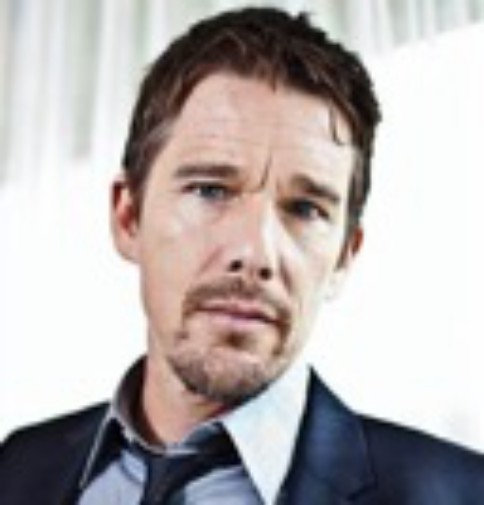} &
\texttt{Ethan Hawke to Star as Jazz Great Chet Baker in New Biopic} \\
\midrule
\includegraphics[width=2.5cm]{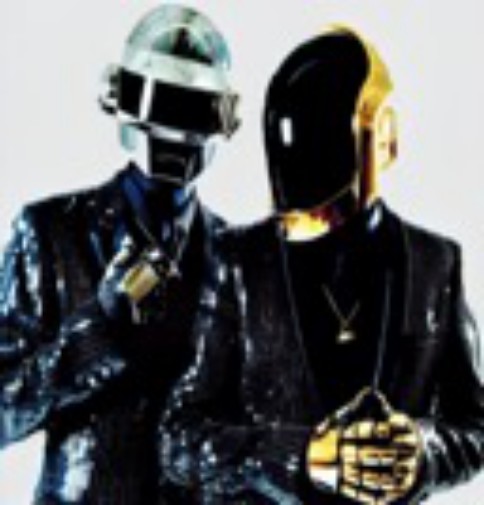} &
\texttt{Daft Punk, Jay Z Collaborate on Computerized} \\
\midrule

\includegraphics[width=2.5cm]{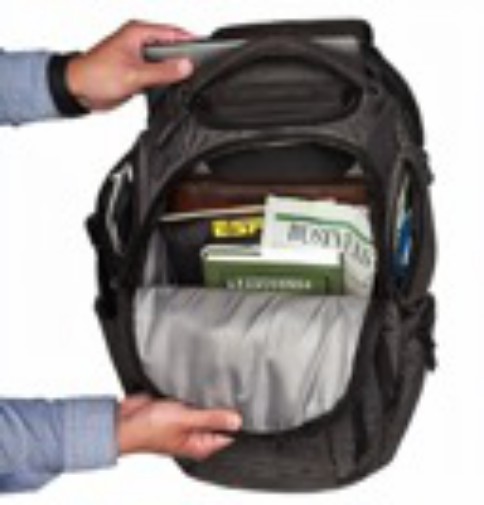} &
\texttt{Renegade RSS Laptop Backpack - View 4} \\
\midrule

\includegraphics[width=2.5cm]{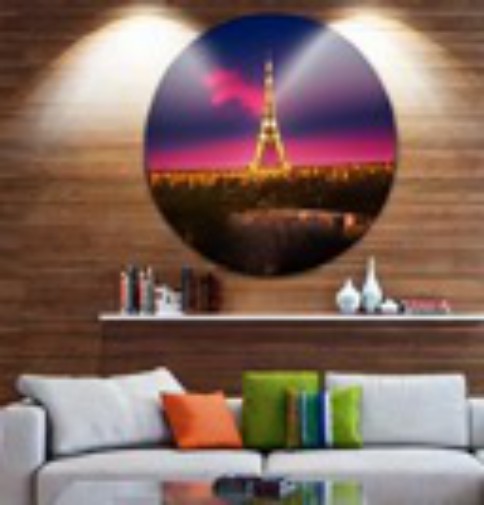} &
\texttt{Design Art Beautiful View of Paris Eiffel Towerunder Red Sky Ultra Glossy Cityscape Circle Wall Art} \\
\midrule

\includegraphics[width=2.5cm]{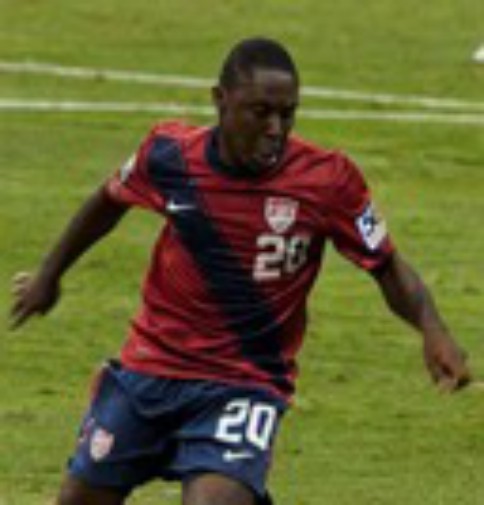} &
\texttt{Freddy Adu Signs For Yet Another Club You Probably Don't Know} \\
\midrule

\includegraphics[width=2.5cm]{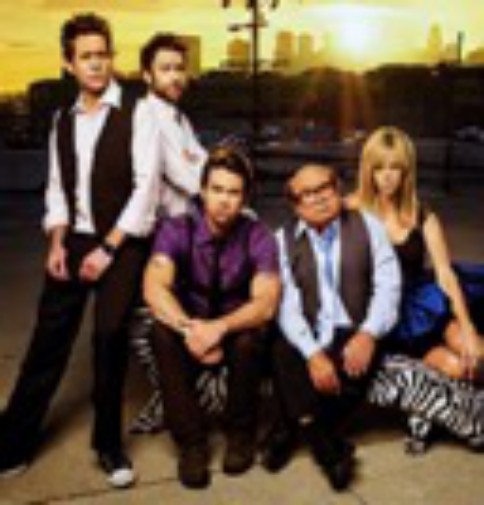} &
\texttt{It's Always Sunny Gang Will Turn Your Life Around with Self-Help Book} \\
\midrule

\includegraphics[width=2.5cm]{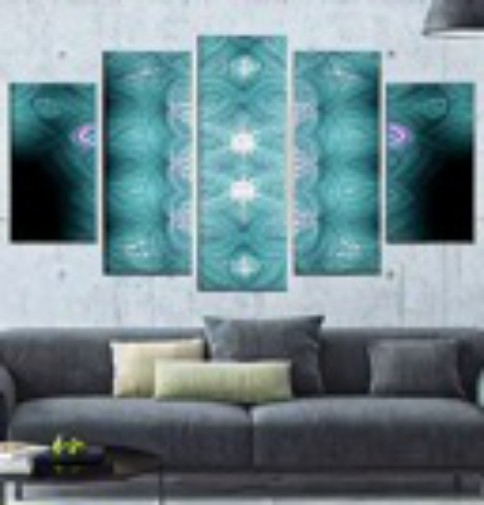} &
\texttt{Designart Circled Blue Psychedelic Texture Abstract Art On Canvas - 7 Panels} \\
\midrule

\includegraphics[width=2.5cm]{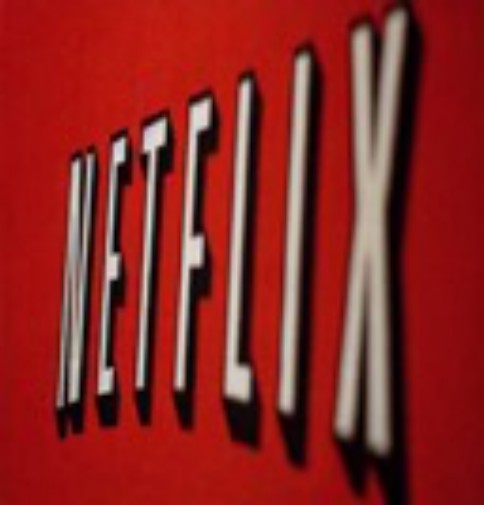} &
\texttt{Netflix Strikes Deal with AT\&T for Faster Streaming} \\
\midrule

\includegraphics[width=2.5cm]{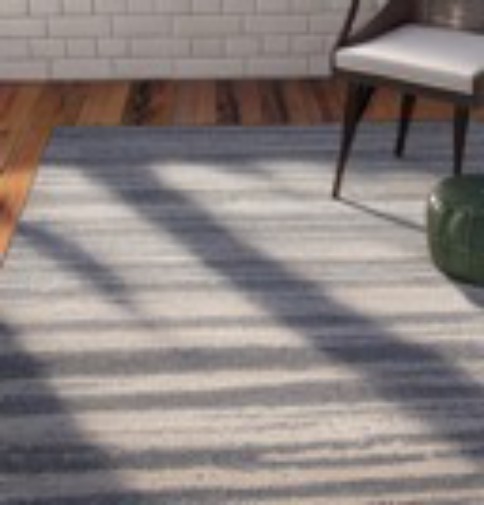} &
\texttt{Shaw Floors Sandy Hollow III 15' Adobe 00108\_Q4278} \\
\midrule

\includegraphics[width=2.5cm]{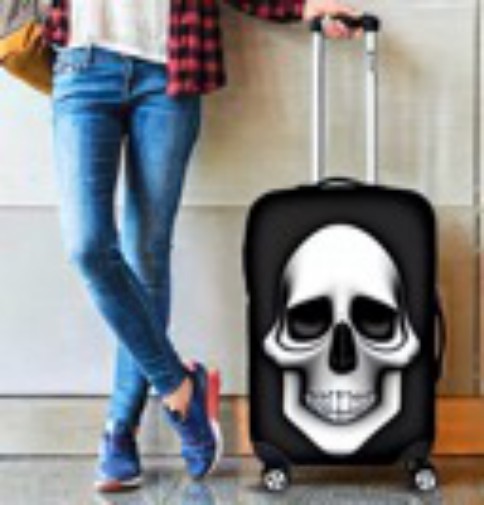} &
\texttt{3D Black \& White Skull King Design Luggage Covers 007} \\
\bottomrule

\end{tabular}}
\end{subtable}
\end{table*}

\section{Additional Qualitative Results}

We provide additional qualitative results in \cref{fig:supp_qualitative1} and \cref{fig:supp_qualitative2}, along with the corresponding prompts listed in \cref{tab:prompts}. \Cref{tab:prompts} also includes the prompts used in \cref{fig:main_qualitative} of the main paper. As shown in the figures, our approach achieves structural changes while maintaining strong semantic alignment, a property that methods like BE~\cite{chen2025exploring} often struggle with. PRSS~\cite{chen2025enhancing} performs better in this regard but can deviate from the intended prompt due to the semantic prompt replacement. ~\cite{hanadjusting} can also induce structural changes through its initialization strategy, which aligns with our frequency-decomposed noise initialization, allowing early latent steps to incorporate external structural bias before memorization occurs. However, \cite{hanadjusting} sometimes leads to over-saturated visual effects, likely due to its proposed noise-initialization optimization approach. Regarding fine-grained details, we observe that direct feature injection can shift the denoising trajectory, leading some visual details to drift away from the memorized image. At the same time, this shift also strengthens the correctness of textual representation in the final output compared to other methods.



\section{Notation}
\begin{table*}[t]
\centering
\renewcommand{\arraystretch}{1.22}
\caption{Main notation used in our paper.}
\label{tab:notation}
\resizebox{0.95\linewidth}{!}{
\begin{tabular}{llp{10.5cm}}
\toprule
\multicolumn{2}{l}{\textbf{Diffusion Process Variables}} \\ 
\midrule
$\mathbf{x}_t$ & latent variable at timestep $t$ &
Represents the noisy latent during denoising; evolves from $\mathbf{x}_T \sim \mathcal{N}(0,\mathbf{I})$ to final latent $\mathbf{x}_0$. \\

$\mathbf{x}_0$ & clean latent (final) &
Predicted clean latent after denoising, later decoded to an image. \\

$\hat{\mathbf{x}}_0$ & predicted clean latent &
Computed using the denoising network $\epsilon_\theta$ (Eq.~\ref{eq:predicted_x0}). \\

$\hat{\mathbf{x}}'_0$ & injected clean latent &
Clean latent after feature injection inside the mask region. \\

$\epsilon, \epsilon_t, \epsilon_T$ & Gaussian noises &
$\epsilon$ is forward‐diffusion noise; $\epsilon_t$ is reverse‐step noise; $\epsilon_T$ is initial noise for sampling. \\

$\beta_t$ & variance schedule &
Defines forward diffusion variance. \\

$\alpha_t = 1-\beta_t$, $\bar{\alpha}_t$ & noise schedule terms &
Used to compute the reverse mean and predicted clean latents. \\

$\epsilon_\theta(\cdot)$ & noise prediction network &
U-Net noise estimator conditioned on text. \\

$E, D$ & VAE encoder/decoder & 
Used to map images to latents and back. \\

$c$ & text-conditioning embedding &
Encoding of the prompt used by the diffusion model. \\
\midrule

\multicolumn{2}{l}{\textbf{Reference Image Retrieval and Scoring}} \\
\midrule
$\mathbf{p}$ & input prompt & Text description used for generation. \\

$w_i,\, w^{*}$ & top-$k$ attended words &  
Words extracted by cross-attention aggregation; $w^{*}$ is randomly sampled for retrieval. \\

$\mathcal{X}_{\text{web}}$ & retrieved image set &  
Candidate images from Pexels/Unsplash. \\

$\mathbf{X}_j$ & $j$th candidate image &
A retrieved non-memorized image. \\

$\mathbf{X}_r$ & selected reference image &
Final reference image chosen after scoring. \\

$V(\cdot), T(\cdot)$ & CLIP vision/text encoders &
Used to compute semantic similarity. \\

$\mathbf{f}_j$ & CLIP embedding of $\mathbf{X}_j$ &
Vision features used in retrieval scoring. \\

$\mathbf{g}^{*}$ & CLIP embedding of $w^{*}$ &
Query word embedding. \\

$h_{1}, h_{2}, h_{3}$ & scoring terms &
Semantic alignment, dataset novelty, and perceptual uniqueness scores. \\

$\boldsymbol{\lambda}$ & scoring weights &
Fixed weights $[0.3,\,0.4,\,0.3]$ for retrieval scoring. \\
\midrule

\multicolumn{2}{l}{\textbf{Frequency Decomposition Initialization}} \\
\midrule
$\mathbf{x}_T$ & initialization latent &
Constructed from high-frequency noise and low-frequency reference features. \\

$\mathbf{x}_r$ & latent of reference image &
Computed as $E(\mathbf{X}_r)$. \\

$\mathcal{F}, \mathcal{F}^{-1}$ & Fourier transform / inverse &
Used to blend high/low frequencies. \\

$\mathcal{M}_{\text{low}},\mathcal{M}_{\text{high}}$ & frequency masks &
Complementary low/high frequency masks in Fourier domain. \\
\midrule

\multicolumn{2}{l}{\textbf{Timestep Window Localization}} \\
\midrule
$s_t$ & CLIP similarity at timestep $t$ &
Measures semantic alignment of $\mathbf{x}_t$ to prompt $\mathbf{p}$. \\

$t_{\text{low}}, t_{\text{high}}$ & injection window bounds &
Denoising steps where memorization tends to emerge. \\

$\frac{\mathrm{d}s_t}{\mathrm{d}t}$ & similarity derivative &
Used to detect phase transition of semantic refinement. \\

$\mu_{\mathrm{d}s_t/\mathrm{d}t},\, \sigma_{\mathrm{d}s_t/\mathrm{d}t}$ &
mean and std of derivative &
Used to threshold the lower bound of injection window. \\
\midrule

\multicolumn{2}{l}{\textbf{Spatial Memorization Localization}} \\
\midrule
$\mathbf{m}_{\text{BE}}$ & Bright Ending mask &
Localizes regions correlated with memorized content. \\

$\mathbf{m}_{\text{concept}}$ & concept-attention mask &
Refined spatial mask aligned with the target concept. \\

$\mathbf{m}$ & final binary mask &
Intersection mask: $\mathbf{m} = \mathbf{1}_{>\tau}(\mathbf{m}_{\text{BE}} \odot \mathbf{m}_{\text{concept}})$. \\

$\tau$ & threshold &
Used to binarize mask intersection. \\
\midrule

\multicolumn{2}{l}{\textbf{Feature Injection}} \\
\midrule
$\delta$ & injection strength &
Controls the intensity of reference feature injection. \\

$\odot$ & elementwise multiplication &
Used for mask operations and blending. \\
\bottomrule
\end{tabular}%
}
\end{table*}

For clarity and completeness, we provide in \cref{tab:notation} a summary of the notation used throughout the main paper. The table groups variables by their functional role in the method, including diffusion-process variables, reference-image retrieval and scoring, frequency-decomposition initialization, timestep-window localization, spatial-memorization localization, and feature injection. This overview is intended to help readers follow the mathematical definitions and implementation details presented in the accompanying sections.

\end{document}